%% file: acl_latex.tex
\documentclass[11pt]{article}

\usepackage[final]{acl}

\usepackage{times}
\usepackage{latexsym}

\usepackage[T1]{fontenc}

\usepackage[utf8]{inputenc}

\usepackage{microtype}

\usepackage{inconsolata}

\usepackage{graphicx}

\usepackage{booktabs}
\usepackage{multirow}
\usepackage{fdsymbol}
\usepackage[most]{tcolorbox}
\usepackage{kotex}
\usepackage{CJKutf8}
\usepackage{xspace}
\usepackage{enumitem}
\usepackage{pifont}
\usepackage[dvipsnames]{xcolor}
\usepackage[normalem]{ulem}
\usepackage{expex}
\usepackage{subcaption}
\usepackage{fvextra}
\def\eg{\emph{e.g}.,\xspace}
\def\ie{\emph{i.e}.,\xspace}
\def\framework{\textbf{\texttt{CuCu}}\xspace}
\def\data{\textbf{\texttt{KCaQA}}\xspace}

\useunder{\uline}{\ul}{}

\title{From National Curricula to Cultural Awareness: \\Constructing Open-Ended Culture-Specific Question Answering Dataset}

\author{
 \textbf{Haneul Yoo$^\heartsuit$} \hspace{5mm}
 \textbf{Won Ik Cho$^\clubsuit$\setcounter{footnote}{1}\Thanks{Work done independently.}} \hspace{5mm}
 \textbf{Geunhye Kim$^\diamondsuit$} \hspace{5mm}
 \textbf{Jiyoon Han$^\spadesuit$}
\\
\\
 $^\heartsuit$New York University \quad
 $^\clubsuit$AI Center, Samsung Electronics \\
 $^\diamondsuit$Hankuk University of Foreign Studies \quad
 $^\spadesuit$Upstage
\\
\\
 \texttt{
   \href{mailto:haneul.yoo@kaist.ac.kr}{haneul.yoo@nyu.edu}
 }
}

\begin{document}
\maketitle

\begin{abstract}
\input{texes/0_abstract}
\end{abstract}

\section{Introduction}
\input{texes/1_introduction}

\section{Related Work}
\input{texes/2_related_work}

\section{\framework and \data}
\input{texes/3_framework}

\section{Discussions}
\input{texes/4_discussion}

\section{Conclusion}
\input{texes/5_conclusion}

\section*{Limitation}
\input{texes/limitation}

\section*{Ethical Considerations}
\input{texes/ethical_considerations}

\section*{Acknowledgments}
This work was conducted as part of the Sovereign AI Foundation Model Project (Data Track), organized by MSIT and supported by the National Information Society Agency (NIA), South Korea (Grant No. 2026-AIData-WII03). In addition, this research was supported by the MSIT, South Korea, under the Top-Tier AI Global HRD invitation program (RS-2025-25461932) supervised by the IITP
(Institute for Information \& Communications Technology Planning \& Evaluation).
We use ChatGPT, Gemini, Claude, and Copilot for writing and coding assistance.

\bibliography{custom}

\clearpage
\appendix
\section*{Appendix}
\input{texes/appendix}

\end{document}

%% file: texes/0_abstract.tex
Large language models (LLMs) achieve strong performance on many tasks, but their progress remains uneven across languages and cultures, often reflecting values latent in English-centric training data.
To enable practical cultural alignment, we propose a scalable approach that leverages national social studies curricula as a foundation for culture-aware supervision.
We introduce \framework, an automated multi-agent LLM framework that transforms national textbook curricula into open-ended, culture-specific question–answer pairs for supervised fine-tuning (SFT).
Applying \framework to the Korean national social studies curriculum, we construct \data, comprising 34.1k open-ended QA pairs.
Our analyses and training experiments suggest that \data covers culture-specific topics and produces responses grounded in local sociocultural contexts.\thinspace\footnote{\framework and \data are available at \\ \url{https://github.com/haneul-yoo/cucu}.}

%% file: texes/1_introduction.tex
Large language models (LLMs) have demonstrated remarkable performance on complex tasks, yet this progress has been uneven across languages and cultures.
As training corpora are dominated by English-centric resources~\cite{qin2025survey, zhu2024multilingual, blasi-etal-2022-systematic}, LLMs often internalize values and cultural framing latent in these resources.
This yields culturally hollow outputs or inappropriate behaviors when applied to underrepresented regions~\cite{tao2024cultural, alkhamissi-etal-2024-investigating, held2023material}.
As LLMs are deployed in culture-sensitive domains, this skew becomes consequential, leading to reduced performance, biased judgments, and limited cultural sensitivity in local sociocultural contexts~\cite{romanou2025include, kwok2024evaluating, liu-etal-2024-multilingual, foroutan-etal-2022-discovering}.

Post-training (\eg supervised fine-tuning and preference optimization) offers a practical path toward culture-adapted or sovereign LLMs~\cite{xu-etal-2025-self, guo-etal-2025-care, masoud-etal-2025-cultural}.
However, obtaining high-quality supervision grounded in local civic norms, historical narratives, and social institutions remains a major bottleneck~\cite{laiyk-etal-2025-instruction}.
Existing multilingual post-training datasets are typically general-purpose (\eg daily life, STEM, reasoning) and often produced via translation of English resources~\cite{li2023bactrianx}, which provides linguistic coverage but weak cultural grounding~\cite{rachamalla-etal-2025-pragyaan}.

In this paper, we propose using national curricula as a structured prior for generating culture-grounded supervision at scale.
Unlike ad hoc prompts or translated datasets, national curricula are carefully designed by domain experts and, in many countries, publicly published by governments or educational authorities; they explicitly enumerate what students are expected to learn and thereby provide both coverage (which topics a society prioritizes) and constraints (how those topics should be framed in relation to civic norms and institutions).
This structure offers a principled backbone for constructing open-ended, culture-specific question answering for supervised fine-tuning (SFT).

\input{sources/figure/framework}

We introduce \framework (from national \textbf{Cu}rricula to \textbf{Cu}ltural Awareness), a multi-agent LLM framework that leverages textbook curricula to automatically generate open-ended, culture-specific question–answer pairs (Figure~\ref{fig:framework}).
We apply \framework to the Korean national social studies curriculum to produce \data (\textbf{K}orean \textbf{C}ulturally-\textbf{a}ware \textbf{QA}), comprising 34.1k open-ended, culture-specific QA pairs on Korean culture in four languages.
By grounding the generation process in national standards, \framework ensures that the resulting data incorporates culture-specific knowledge as well as normative perspectives required for genuine cultural awareness.

%% file: sources/figure/framework.tex
\begin{figure*}[h]
    \centering
    \includegraphics[width=\linewidth]{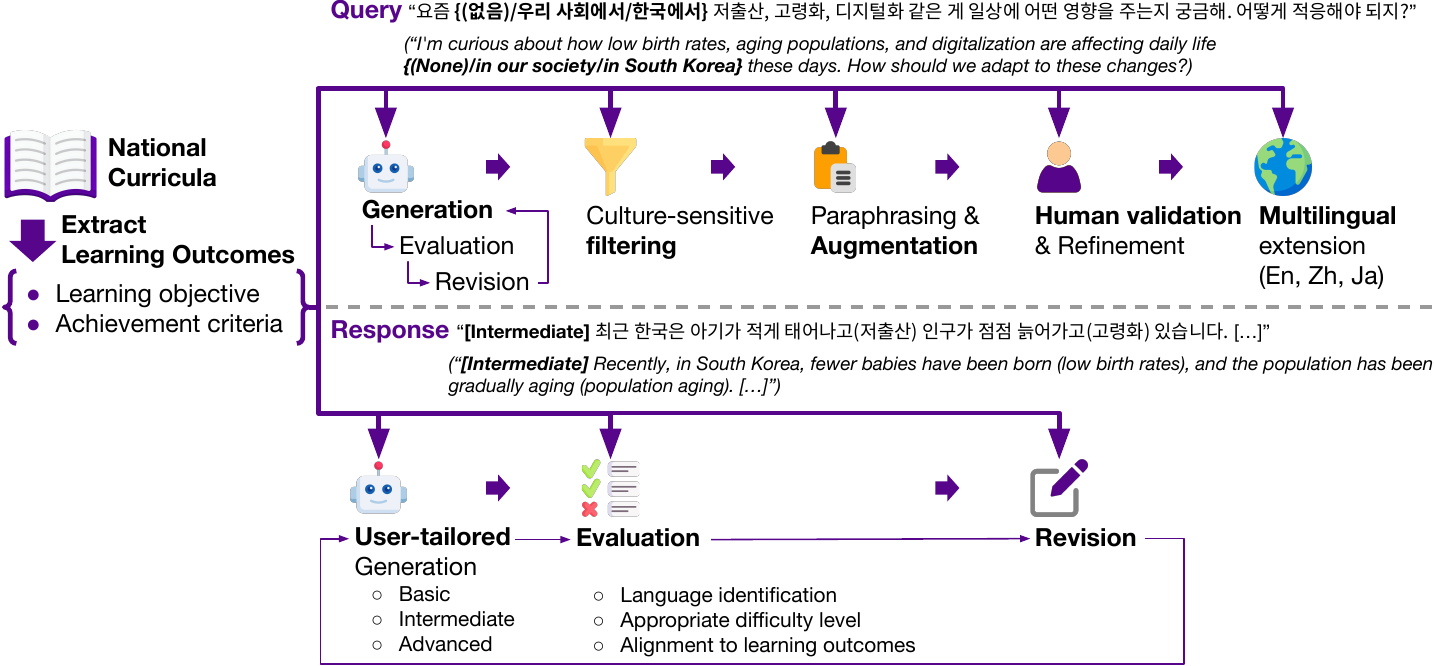}
    \caption{Overview of \framework. \framework constructs open-ended culture-specific QA pairs using national curricula.}
    \label{fig:framework}
\end{figure*}

%% file: texes/2_related_work.tex
\subsection{AI-assisted Data Construction}

The construction of natural language data has traditionally been conducted manually, but the recent disclosure of generative language models has enabled practitioners to obtain such resources with the help of machines.
Especially for the pre-training and post-training data of language models, \citet{gunasekar2023textbooks} has demonstrated the significance of synthetic textbook-like materials as a sufficient and necessary source of learning.
In this light, Cosmopedia~\cite{benallal2024cosmopedia} investigated diverse depths in data synthesis, considering a wide range of domains and audiences. 
These approaches led to multiple data generation attempts in alignment~\cite{wang2024codeclm}, reducing hallucination~\cite{jones2024teaching}, and further synthesizing frameworks~\cite{maini2025beyondweb}.

\subsection{Culture-specific NLP Data}

Culture-specific NLP data has long been a topic of interest among diversity and inclusion interest groups. 
While culture incorporates a wide range of dimensions~\cite{spencer2012culture}, prior works have primarily focused on ensuring or benchmarking language models' capability to understand professional knowledge, such as customs~\cite{li2024culturepark, myung2024blend, chiu-etal-2025-culturalbench}, norms~\cite{rao2025normad}, perceptions~\cite{shi-etal-2024-culturebank}, and common grounds~\cite{tanwar2025you} shared among cultural groups.
Furthermore, these data primarily consist of closed-ended and short-form QAs, with human-authored and validated items, which differ from natural user queries and make scaling challenging.

Especially for Korean, prior studies have evaluated cultural alignment as specific tasks, focusing on cultural heritage~\cite{kim2022kochet}, historical knowledge~\cite{son2024hae}, factuality~\cite{ko2025kosimpleqa}, or social bias~\cite{jin2024kobbq, lee-etal-2023-kosbi}. 
While \citet{lee2024kornat} investigated social values and common knowledge regarding Korean among LLMs, the queries are closed-ended and does not necessarily account for LLM-targeted questions.
In this light, we aim to introduce a scalable framework and the following dataset to address open-ended, culture-specific questions based on social consensus grounded in national curricula.

%% file: texes/3_framework.tex
We introduce \framework (from national \textbf{Cu}rricula to \textbf{Cu}ltural Awareness), a multi-agent LLM framework that constructs an open-ended, culture-specific  QA dataset using national curricula.
Figure~\ref{fig:framework} describes an overview pipeline of \framework.
Details on \framework are described in Appendices~\ref{sec:framework_details} and \ref{sec:system_prompt}.
Our rationale for the design choices of \framework, including model selection, is described in Appendix~\ref{sec:design}.
The following steps in \framework are sequential.
For all internal evaluations, we use a 1–5 Likert scale and let models revise and regenerate data based on the feedback, up to 5 times if the scores are below 5.

\paragraph{Query Generation.}
We use \texttt{Solar Pro 2}~\cite{upstage2025solarpro2} and \texttt{GPT-4o}~\cite{openai2024gpt4} to generate or revise query texts and to evaluate query texts in each LLM agent, respectively.
Experimental results of replacing and reducing the reliance on the proprietary model (\ie \texttt{GPT-4o}) are described in Appendix~\ref{sec:kcaqa_open}.

\begin{enumerate}[leftmargin=*]
    \setlength\itemsep{-0.2em}
    \item \textbf{Initial query generation:} We generate an initial query in Korean based on a learning outcome (\ie a pair of a learning objective and an achievement criterion). We (1) generate a query, (2) evaluate whether it fully reflects the learning outcome, and (3) revise based on the feedback, repeating this loop up to five times.
    \item \textbf{Culture-sensitive filtering:} We filter out queries that involve general, culture-agnostic knowledge, further improving their naturalness and language use.
    \item \textbf{Paraphrasing \& Augmentation:} We paraphrase the initial query into two additional queries that target the same learning outcomes but differ in style and tone. Each resulting query is expanded into three variants: (1) without mentioning any culture or country, (2) implicitly referring to ``our/my'' culture or country, and (3) explicitly stating the name of culture or country.
    \item \textbf{Human validation \& Refinement:} One of the authors, who is a Korean L1 speaker born and raised in Korea, educated under the Korean national curricula and thus familiar with its norms and culture, manually validates and refines queries generated by multi-agent LLM.
    \item \textbf{Multilingual extension:} We finally extend the queries into three additional languages (\ie English, Chinese, and Japanese) by translating the ones explicitly stating the culture/country name to avoid ambiguity.
\end{enumerate}

\paragraph{Response Generation.}
We use four open-source LLMs (\ie \texttt{GPT-OSS}~\cite{openai2025gptoss}, \texttt{Qwen3}, \texttt{Qwen3-Next}~\cite{yang2025qwen3}, and \texttt{DeepSeek R1}~\cite{deepseekai2025deepseekr1}) for response.

\begin{enumerate}[leftmargin=*]
    \setlength\itemsep{-0.2em}
    \item \textbf{User-tailored response generation:} For each query, we produce three variants targeting different audience bands---\emph{Basic}, \emph{Intermediate}, and \emph{Advanced}---to cover a range of readability and assumed background knowledge.
    \item \textbf{Response evaluation:} We evaluate each candidate for (1) language consistency with the query, (2) appropriateness of difficulty in terms of style, readability, and depth, and (3) alignment with the corresponding learning outcomes.
    \item \textbf{Response revision:} We revise the response and repeat this process up to five times.
\end{enumerate}

To showcase \framework, we use the 2022 Revised Elementary, Middle, and High School National Curriculum of Social Studies\thinspace\footnote{\url{https://ncic.re.kr/}}, published by the Ministry of Education, Republic of Korea, as a source data.
We extract 357 atomic pairs of learning objectives and achievement criteria and filter out 158 learning outcomes, generating 2,844 queries (\ie 3 paraphrasing $\times$ (3 Korean variants $+$ 3 translations) $\times$ 158 learning outcomes).
In total, we construct \data (\textbf{K}orean \textbf{C}ulturally-\textbf{a}ware \textbf{QA}), comprising 34,128 QA pairs (\ie 2,844 queries $\times$ 3 difficulty levels $\times$ 4 models) that reflect Korean cultural knowledge and perspective. 
An example instance of \data is provided in Appendix~\ref{sec:data_samples}.
We provide further details on our design choices and discuss wider adaptability in Appendices~\ref{sec:design} and~\ref{sec:adapt}.
We also discuss potential use cases for \data in Appendix~\ref{sec:use_case}.

%% file: texes/4_discussion.tex
\subsection{\data Quantitative Analysis}
\label{sec:quant_analysis}

\paragraph{Culture-specific Query.}

We analyze whether \data queries cover culture-specific topics using topic modeling.
In particular, we compare the Korean explicit queries in \data with Korean LIMA~\cite{zhou2023lima} dataset\thinspace\footnote{Translated version of LIMA dataset. Accessible at \url{https://huggingface.co/datasets/taeshahn/ko-lima}}, a high-quality, general-domain QA dataset for LLM alignment training.
The resulting distributions diverge sharply, with a Jensen–Shannon Divergence (JSD) of 0.807 between \data and Korean LIMA; a 1k permutation test confirms the difference is significant ($p<0.01$).
A qualitative inspection of the top-10 most skewed topics suggests that \data mainly covers Korean sociocultural content (\eg Korean geography, economy, history, climate, and politics).

\paragraph{User-tailored Response.}
We examine whether \data responses match their intended difficulty level (\ie \emph{Basic}, \emph{Intermediate}, and \emph{Advanced}) using readability statistics, as reported in Table~\ref{tab:readability}.
In general, \data responses become longer and syntactically denser with increasing target level, while lexical diversity also increases.
Further analyses on both query and response of \data are provided in Appendix~\ref{sec:appendix_quant_analysis}.

\input{sources/table/training}

\subsection{\data Quality Estimation}
\label{sec:quality_estimation}

We validate the data quality of \data by LLM-as-a-judge~\cite{zhent2023judging} and human inspection.
The evaluation criteria with detailed rubrics are described in Appendix~\ref{sec:rubric}.

\begin{itemize}[leftmargin=*]
    \setlength\itemsep{-0.2em}
    \item \textbf{Language selection}: If both query and response are monolingual in the target language;
    \item \textbf{Cultural appropriateness}: If both query and response adhere to the desired stance regarding culture-sensitive concepts;
    \item \textbf{Language use}: If both query and response use appropriate and natural style and expressions.
\end{itemize}

We employ \texttt{GPT-5.2}~\cite{openai2025gpt5} to automatically assess the quality of \data.
Specifically, the model outputs a binary decision for \emph{language selection} (0/1) and assigns Likert-scale scores (1--10) for \emph{cultural appropriateness} and \emph{language use}.
\data achieves 0.91 accuracy on language selection, with scores of 8.56 and 7.78 for cultural appropriateness and language use, respectively.
These results suggest that \data produced by \framework largely satisfies the intended language constraints and stance requirements for culture-specific concepts.
In addition, we manually inspect all Korean instances and a randomly sampled set of multilingual instances.
We observe that, in most cases, queries capture the core topic of the learning objective, and responses align with the corresponding criteria.
The full results, including human validations and failure analysis, as well as quantitative analysis on multilingual extension, are discussed in Appendix~\ref{sec:llm_as_a_judge}.

\subsection{LLM Training using \data}
\label{sec:training}

We train \texttt{Qwen3-14B} on five datasets for Korean cultural alignment: (1) CultureBank~\cite{shi-etal-2024-culturebank}, a multilingual, multicultural dataset sourced from Tiktok and Reddit, (2) Korean LIMA, a general-purpose, open-domain QA dataset, and three ablation variants of \framework (\ie (3) national curricula, (4) \framework without source data, and (5) \data).
We use CLIcK~\cite{kim-etal-2024-click} cultural questions, BLEnD~\cite{myung2024blend} multiple-choice questions, CulturalBench~\cite{chiu-etal-2025-culturalbench}-Hard for cultural knowledge evaluation, as well as KorNAT~\cite{lee2024kornat} for cultural alignment evaluation, all tailored for Korean culture.
Experimental settings are described in Appendices~\ref{sec:apendix_training} and \ref{sec:kcaqa_open}.

Table~\ref{tab:training} shows that the model trained with KCaQA yields the strongest gains on both cultural knowledge and alignment.
While training on the raw curriculum improves performance, it is weaker than training on \data.
It implies that converting curricula into structured, open-ended supervision provides non-trivial, transferable gains. Furthermore, \framework without source grounding data also improves cultural knowledge and alignment, indicating that the framework design itself contributes.
However, using CultureBank does not substantially enhance cultural knowledge or alignment, likely because it is sourced from Reddit or TikTok and primarily focuses on how people from other cultures perceive cultural differences and culture shock rather than grounded civic norms.

%% file: sources/table/training.tex
\begin{table*}[htb!]
\centering
\resizebox{0.73\linewidth}{!}{
\begin{tabular}{@{}l|ccc|c@{}}
\toprule
                                                               & CLIcK & BLEnD                      & CulturalBench & KorNAT \\ \midrule
Qwen3-14B                                                     & 0.587 & 0.842 & 0.585              & 0.679  \\
\hspace{2mm} + 1. CultureBank & 0.611 & 0.859                      & 0.607              & 0.655  \\
\hspace{2mm} + 2. Korean LIMA              & 0.615 & 0.853                      & 0.611              & 0.682  \\
\hspace{2mm} + 3. National curriculum                          & 0.633 & 0.866                      & 0.615              & 0.721  \\
\hspace{2mm} + 4. \framework without source data               & 0.636 & 0.870                      & 0.624              & 0.736  \\
\hspace{2mm} + 5. \data (Ours)                                 & \textbf{0.662} & \textbf{0.891}                      & \textbf{0.634}              & \textbf{0.744}  \\ \bottomrule
\end{tabular}
}
\caption{Experimental results of LLM post-training (SFT) using \framework and \data tailored for Korean culture. We use CLIcK, BLEnD, and CulturalBench for cultural knowledge evaluation and KorNAT for cultural alignment evaluation. We extract and use the samples targeting Korean culture for BLEnD and CulturalBench.}
\label{tab:training}
\end{table*}

%% file: texes/5_conclusion.tex
We propose a scalable recipe for culture-grounded post-training supervision by leveraging national social studies curricula as an expert-structured prior. 
We introduce \framework, an automated multi-agent LLM pipeline that converts learning outcomes into open-ended, culture-specific QA pairs, and apply it to the Korean national social studies curriculum to generate \data comprising 34.1k QA pairs across four languages. 
Through quantitative and qualitative analyses, we show that \data captures curriculum-specified, culture-specific topics and yields answers that are grounded in Korean sociocultural context across languages.
Through controlled SFT experiments, we empirically validate the effectiveness of \framework and \data for cultural alignment.
This study provides a practical and reproducible path to constructing culturally grounded supervision for sovereign or culture-adapted LLMs.

%% file: texes/limitation.tex
This study incorporates several conceptual and methodological limitations.

First, the concepts of language, linguistic region, cultural region, ethnicity, and nation-state each carry distinct meanings, yet our approach examines cultural reflection primarily through the lens of national curricula. This methodology proves feasible for South Korea, where its compact nation-state highly aligns with the cultural boundary. However, this approach faces challenges when applied to nation-states in multiple geocultural regions with varying cultural contexts.

Second, our analysis focuses exclusively on Seoul Korean as used in the Republic of Korea (ROK) and references its standard curriculum. This scope excludes Pyongyang Korean, as employed in the Democratic People's Republic of Korea (DPRK), and does not consider North Korea's educational standards. While we prioritize learning outcomes from the Korean curriculum that focus on widely accepted civic and social knowledge, certain aspects may not align with North Korean linguistic or educational norms.

Third, our use of national curricula as primary sources for query and response generation, while ensuring verified and socially accepted content, does not fully capture the range of geocultural characteristics expressed through language. Curricula represent institutionalized knowledge rather than the complete spectrum of cultural expression. Nevertheless, our framework maintains sufficient flexibility to accommodate alternative source materials that may more comprehensively represent linguistic and cultural phenomena when such resources become available.

%% file: texes/ethical_considerations.tex
\paragraph{LLM-generated Data.}

Our \framework mainly consists of automated data construction and validation, which is transparently shared to support reproducibility. We acknowledge, however, that data quality is paramount and that LLM-generated data may introduce model-specific biases, hallucinations, stylistic artifacts, or over-standardized framings.
Our use of LLM-generated data is not intended to replace human judgment, but to address known challenges in manually constructing culture-specific datasets. Human-only generation can suffer from individual annotator bias and limited topical diversity~\cite{geva-etal-2019-modeling}, as annotators inevitably draw on personal, regional, socioeconomic, or ideological backgrounds~\cite{paullada2021data, cui-etal-2025-bias}. By grounding LLM generation in national curricula through carefully designed prompts, our approach enables broader and more standardized coverage of culture-specific knowledge. This choice aligns with recent NLP work on scalable LLM-based annotation and data synthesis~\cite{tan-etal-2024-large}; prior studies further show that LLMs can achieve annotation quality comparable to, or sometimes better than, human crowdworkers while reducing individual-level human variance~\cite{gilardi2023chatgpt}.
Nevertheless, LLM-generated data should not be treated as objective or bias-free. Our methodology may inherit biases from underlying LLMs, including those from pretraining data, alignment procedures, and generation styles. To mitigate these risks, we place human inspection: validating and refining generated queries before multilingual extension. This step helps identify unnatural, overly generic, weakly grounded, or potentially misleading queries while preserving the scalability and topical coverage of LLM-based generation.

\paragraph{Controversy in National Curricula.}
We regard the national curricula of the Republic of Korea as reliable and institutionally grounded data sources, but not as fully neutral or controversy-free. The 2022 revised Korean social studies curriculum reflects institutional consensus and public- and expert-driven revision processes~\cite{cho2023analysis, lee2024presence}, but this does not guarantee that all concerns are resolved or that the curriculum represents a unanimous view.
Accordingly, models trained on \framework, queries, or answers may inherit or amplify curricular biases. This risk is especially important when applying \framework to other national or cultural regions, where curricula may vary across institutions, regions, political systems, or stakeholder groups. As multiple nations continue to debate bias, omission, or revisionism in national curricula, practitioners should carefully examine adopted curricula and recognize that unresolved domestic or international controversies may propagate into generated datasets and downstream models.
To navigate these risks without imposing author-defined filtering rules, we rely on grounded sources. Rather than filtering controversial topics, the generated responses follow the adopted curriculum, avoid unsupported claims, and, when relevant, distinguish established evidence from interpretations. For diplomatic or institutionally defined issues, practitioners should avoid introducing independent geopolitical claims beyond the curriculum-grounded framing. We do not independently adjudicate unresolved controversies, as doing so would require additional subjective judgments and could introduce another source of bias.

%% file: texes/appendix.tex
\section{Reproducibility Statement}
For open-source LLM inference, we use 4 H200 GPUs with 564 GB memory and 4 RTX 8000 GPUs with 188 GB memory.
For proprietary LLM inference, we use the official API implementation of each model.
We use a temperature of 0.0 (\ie greedy decoding) for the generations in \framework.
We specifically use the versions of \texttt{gpt-4o-2024-08-06} and \texttt{gpt-5.2-2025-12-11} for \texttt{GPT-4o} and \texttt{GPT-5.2}, respectively.
We use the following LLMs for response generation via Hugging Face:
\begin{itemize}[leftmargin=*]
    \setlength\itemsep{-0.2em}
    \item \texttt{openai/gpt-oss-120b}, 
    \item \texttt{Qwen/Qwen3-235B-A22B}, 
    \item \texttt{Qwen/Qwen3-Next-80B-A3B-Thinking}, 
    \item \texttt{deepseek-ai/DeepSeek-R1}.
\end{itemize}

In Section~\ref{sec:quant_analysis}, we use Korean Sentence BERT~\cite{kr-sbert} for query analysis to fit a single BERTopic~\cite{grootendorst2022bertopic} model on the two corpora, representing each instance as a document that concatenates the query and its paired response.
For response analysis, tokens are obtained with \texttt{wordfreq}~\cite{robyn2022wordfreq}, a unified multilingual tokenizer.
The rare-token ratio is computed only over in-vocabulary tokens (OOV excluded), using a language-specific threshold set to the bottom 25\% of Zipf-frequency values estimated from all responses in that language.

For the SFT experiments in Section~\ref{sec:training} and Appendices~\ref{sec:apendix_training} and \ref{sec:kcaqa_open}, we use 4 NVIDIA H200 141GB GPUs with CUDA version 12.8 to train and evaluate the models.
We fine-tune \texttt{Qwen3-14B} using Hugging Face libraries for 3 epochs with a batch size of 16, a learning rate of 2e-4, and a warmup ratio of 0.03.
We set the random seed to 42 and disabled gradient clipping.

\section{Implementation Details on \framework}
\label{sec:framework_details}

\subsection{Query Generation}

\paragraph{Initial Query Generation.}
We use \texttt{GPT-4o} as an evaluation agent to assess whether a query (1) reflects the learning objective and (2) incorporates key ideas from the achievement standards. The agent scores alignment on a 5-point Likert scale and provides short revision feedback.

\paragraph{Culture-sensitive Filtering.}
We use \texttt{GPT-4o} as an evaluation agent to filter out queries that involve general, culture-agnostic knowledge according to the following criteria: whether a query (1) aligns with the Korean cultural and social context, (2) reflects social consensus institutionalized in Korean society, (3) is not a generic, universal question, and (4) is meaningfully grounded in the Korean context.
This ensures that queries elicit perspectives shaped by local institutions and normative values, rather than subjective stereotypes. 
For instance, the following questions are rejected and accepted, respectively, through the filtering process:

\begin{tcolorbox}[breakable, enhanced, top=1pt, left=1pt, right=1pt, bottom=1pt, colback=white, title=Example questions within the filtering process]
\begin{itemize}[leftmargin=*]
    \item \textbf{Reject (General/Culture-agnostic):} ``\emph{What are the general causes of colonialism?}''
    \item \textbf{Accept (Korea-specific context):} ``\emph{How did Korean society and culture change during the Japanese colonial period, and what movements did people engage in? Give me a quick rundown of daily life changes and the key players behind the movements.}''
    \begin{itemize}[leftmargin=*]
        \item Reason: The Japanese colonial period is more than a mere historical fact; it is a core narrative prioritized in the curriculum as the foundation of South Korea's constitutional spirit and civic consciousness. 
    \end{itemize}
\end{itemize}
\end{tcolorbox}

\paragraph{Augmentation \& Multilingual Extension.}
We vary the linguistic framing of already culture-grounded queries into three settings: (1) no-country, (2) implicit, and (3) explicit.
The motivation is as follows: even when a user does not explicitly mention the country name``Korea'' in a Korean language query, a culturally aligned model should correctly infer the intended sociocultural context as Korean.
Importantly, the ``no-country'' variant does not remove cultural grounding.
It removes only explicit lexical markers, while the underlying learning outcome and assumed context remain Korean.
This distinction becomes especially important in the multilingual extension stage.
As the query language changes (\eg from Korean to English, Chinese, or Japanese), implicit cultural assumptions tied to Korean no longer hold.
Therefore, in multilingual extension, we translate only the explicitly culture-marked variant, ensuring that the Korean sociocultural grounding remains clear in non-Korean languages.

\paragraph{Human Validation \& Refinement.}
This process mainly includes (1) revising queries to be natural colloquial questions, (2) inspecting whether the variants imply the same meaning regardless of the presence of culture- or country-related expressions, and (3) ensuring that queries adhere to the learning objective and the achievement criteria, not handling topics that are overly generic or too professional.

\section{\data Samples}
\label{sec:data_samples}
\begin{tcolorbox}[breakable, enhanced, top=1pt, left=1pt, right=1pt, bottom=1pt, colback=white, title=Example instance extracted from Korean social studies national curriculum]
\begin{itemize}[leftmargin=*]
    \item Learning objectives: [4사03-01] 최근 사회 변화의 양상과 특징을 파악하고, 그로 인해 나타난 생활모습의 변화를 탐색한다. \\
    (\emph{Translation: [4SA03-01] Identify the patterns and characteristics of recent social changes and explore changes in daily life that have emerged as a result.})
    \item Achievement criteria: [4사03-01]은 사례를 통해 저출산, 고령화, 지능정보화 등 최근 사회 변화가 가져온 현상과 특징을 파악하고, 그로 인해 우리의 생활 모습이 어떻게 변화되는지 분석하도록 설정한 것이다. 이를 통하여 사회 변화에 대응할 수 있는 역량을 기르는 데 초점을 맞추도록 한다. \\
    (\emph{Translation: [4SA03-01] is designed to have students understand the phenomena and characteristics brought about by recent social changes---such as low birth rates, population aging, and intelligent informatization---through examples, and to analyze how these changes affect our daily lives. Through this process, the focus is on fostering the capacity to respond to social changes.})
    \item Query: 
    \begin{itemize}[leftmargin=*]
        \item Korean (No country): 요즘 저출산, 고령화, 디지털화 같은 게 일상에 어떤 영향을 주는지 궁금해. 어떻게 적응해야 되지?
        \item Korean (Implicit): 요즘 우리 사회에서 저출산, 고령화, 디지털화가 일상에 어떤 영향을 주는지 궁금해. 어떻게 적응해야 될까?
        \item Korean (Explicit): 요즘 한국에서 저출산, 고령화, 디지털화가 일상에 어떤 영향을 주는지 궁금해. 어떻게 적응해야 되지?
        \item English: I'm curious about how low birth rates, aging populations, and digitalization are affecting daily life in South Korea these days. How should we adapt to these changes?
        \item Chinese: \begin{CJK*}{UTF8}{gbsn}最近好奇韩国的低出生率， 老龄化和数字化对日常生活有什么影响？ 应该怎么适应这些变化呢？\end{CJK*}
        \item Japanese: \begin{CJK*}{UTF8}{min}最近， 韓国の低出生率， 高齢化， デジタル化が日常生活にどんな影響を与えているのか気になります。 こうした変化にどう適応していけばいいですか？\end{CJK*}
    \end{itemize}
    \item Response:
    \begin{itemize}[leftmargin=*]
        \item Basic: 우리나라에 아기들이 적어지고 어른이 많아졌어요. 그래서 가족이 함께 시간을 보내는 게 더 중요해요. 스마트폰과 컴퓨터를 잘 사용하면 생활이 편해져요. 모두가 서로 도와가며 살아가면 좋아요. \\
        (\emph{Translation: In our country, there are fewer babies being born and more adults. So it’s more important for families to spend time together. If we use smartphones and computers well, life becomes more convenient. It’s good when everyone lives by helping each other.})
        \item Intermediate: 최근 한국은 아기가 적게 태어나고(저출산) 인구가 점점 늙어가고(고령화) 있습니다. 동시에 스마트폰, 인터넷 등 디지털 기술이 일상 곳곳에 스며들어(디지털화) 생활이 편리해졌지만 새로운 적응이 필요합니다. 작은 가정에서는 아이를 돌볼 사람이 부족해 보육 서비스 이용이 늘고, 많은 가정이 조부모와 함께 사는 다세대 형태가 늘어납니다. 디지털화 덕분에 온라인 쇼핑, 원격 근무, 비대면 의료 등 새로운 생활 방식이 보편화되었습니다. 이런 변화에 적응하려면 디지털 기기 사용법을 배우고, 세대 간에 서로 도와주는 문화가 중요합니다. \\
        (\emph{Translation: Recently, Korea has seen fewer babies being born (low birth rate) and the population is gradually getting older (aging). At the same time, digital technologies such as smartphones and the internet have become part of everyday life (digitalization), making life more convenient but also requiring people to adapt in new ways. In smaller households, there often aren’t enough caregivers for children, so the use of childcare services is increasing, and more families are living in multi-generational households with grandparents. Thanks to digitalization, new lifestyles such as online shopping, remote work, and contactless healthcare have become widespread. To adapt to these changes, it is important to learn how to use digital devices and to build a culture where different generations support one another.})
        \item Advanced: 2023년 우리나라 합계출산율은 0.78명으로 세계 최저 수준이며, 65세 이상 인구 비중은 16.5\%에 달해 고령화가 급속히 진행되고 있습니다. 동시에 인터넷 보급률은 96\%에 육박하고, 인공지능·빅데이터 기반 서비스가 일상에 스며들면서 디지털 전환이 가속화되고 있습니다. 이러한 인구·기술 변화는 노동시장 구조조정, 연금·보건복지 재정 압박, 소비 패턴의 세대별 차별화 등을 초래하고 있습니다. 대응 방안으로는 평생학습 체계 강화와 디지털 격차 해소를 위한 국가·지방 차원의 교육 프로그램 확대가 필요합니다. 또한, 세대 간 교류를 촉진하는 커뮤니티 센터와 스마트 시티 인프라를 활용해 고령층의 사회참여와 안전을 지원해야 합니다. 기업은 유연근무제와 가족친화 정책을 도입해 출산·육아 부담을 경감하고, 정부는 출산 장려금·보육 서비스 질 향상을 통해 저출산 문제를 완화할 수 있습니다. 이러한 다각적 접근이 개인과 사회가 함께 변화에 적응하는 기반이 될 것입니다. \\
        (\emph{Translation: In 2023, Korea’s total fertility rate was 0.78, one of the lowest in the world, and the share of the population aged 65 and older reached 16.5\%, showing that aging is progressing rapidly. At the same time, internet penetration is close to 96\%, and digital transformation is accelerating as services based on artificial intelligence and big data spread into daily life. These demographic and technological shifts are causing labor-market restructuring, increasing financial pressure on pensions and health/welfare systems, and widening generational differences in consumption patterns. As responses, it is necessary to strengthen lifelong learning systems and expand education programs at the national and local levels to reduce the digital divide. In addition, community centers that promote intergenerational exchange, along with smart-city infrastructure, should be used to support older adults’ social participation and safety. Companies can reduce the burdens of childbirth and childcare by introducing flexible work arrangements and family-friendly policies, while the government can ease the low birth rate problem through childbirth incentives and improving the quality of childcare services. This multifaceted approach will become the foundation for individuals and society to adapt to change together.})
    \end{itemize}
\end{itemize}
\end{tcolorbox}

\section{Additional Discussions}

\subsection{Design Choices of \framework and \data}
\label{sec:design}

\paragraph{\framework.}
During the response generation in \framework, we produce three response variants---Basic, Intermediate, and Advanced---based on the following dimensions.

\begin{tcolorbox}[breakable, enhanced, top=1pt, left=1pt, right=1pt, bottom=1pt, colback=white, title=Dimensions for three response variants]
\begin{itemize}[leftmargin=*]
    \item \textbf{Basic:} 
    \begin{itemize}[leftmargin=*]
        \item Audience: elementary school students and young children
        \item Language: very simple words, short sentences, and concrete ideas
        \item Style: use everyday examples and simple comparisons; one idea per sentence
        \item Content: provide only the most basic and essential meaning of the topic
        \item Length: around 2--3 short sentences
    \end{itemize}
    \item \textbf{Intermediate:} 
    \begin{itemize}[leftmargin=*]
        \item Audience: high school students and young adults
        \item Language: clear, structured explanations similar to the national curriculum and textbooks
        \item Style: show balanced reasoning and connect personal experience with broader social context
        \item Content: when relevant, incorporate the Learning Objective and Achievement Standards
        \item Length: around 4–6 sentences
    \end{itemize}
    \item \textbf{Advanced:} 
    \begin{itemize}[leftmargin=*]
        \item Audience: adults, university students, graduate-level learners, and professionals
        \item Language: analytical, evidence-based, and conceptually deeper
        \item Style: culturally grounded perspectives common among well-educated Korean adults
        \item Content: offer contextual analysis, balanced arguments, and practical implications; when relevant, mention history, institutions, data, or research
        \item Length: around 6–12 sentences, possibly in two short paragraphs
    \end{itemize}
\end{itemize}
\end{tcolorbox}

For query generation, we intentionally separate the model that generates or revises query texts from the one that evaluates them to reduce its bias toward its own style.
Specifically, we use \texttt{Solar Pro 2}, a Korean-centric frontier LLM under the MIT license, for query generation and revision, and \texttt{GPT-4o}, a strong proprietary multilingual model, for query evaluation.
These choices allow us to cover both general multilingual capability and Korean-specific cultural awareness.
For response generation, we use open-source LLMs (\ie \texttt{gpt-oss-120b}, \texttt{Qwen3-235B-A22B}, \texttt{Qwen3-Next-80B-A3B-Thinking}, and \texttt{DeepSeek -R1}) with Apache or MIT licenses, which have no restrictions on output usage, as the generated data may be used for further model training.
We aim to use diverse models for query and response generation to reflect the different writing styles of models.

Our motivation for the multilingual extension in \framework is to (1) enable models to respond to Korean culture in multiple languages and (2) support sovereign or nationally aligned AI systems that must handle multilingual user inputs while reflecting Korean standards and perspectives.
Here, we decide to extend the queries that explicitly mention culture/countries to avoid ambiguity.
This provides multilingual renderings of Korean culture-grounded supervision, which is hardly available in natural corpora.
We select English, Chinese, and Japanese to reflect the practical needs for building sovereign AI in Korea, as these are the second languages most commonly used in Korea.

\paragraph{\data.}
The up-to-date Korea national curricula released in 2022 incorporate the following subjects: \emph{Korean language}, \emph{Mathematics}, \emph{Social Studies}, \emph{Science}, \emph{Moral Education}, \emph{Technology \& Home Economics}, \emph{Music}, \emph{Fine Arts}, \emph{Physical Education}, \emph{Computer \& Information}, \emph{Classical Chinese}, \emph{Second Foreign Language}, and \emph{Creative Experiential Activities}.
Among them, we use the \emph{Social Studies} curriculum only to build \data, as other subjects largely cover culture-agnostic, general knowledge.
For example, \emph{Moral Education} primarily does not target Korean normative values but rather offers general social philosophy and theory.
In contrast, \emph{Social Studies} encompasses various sub-contents closely related to Korean culture: Korean/Asian/World History, Korean/International Geography, Society \& Culture, Common Social Studies, Politics, Economics, Law \& Society, and Understanding International Relations.

\subsection{Adaptability of \framework and \data}
\label{sec:adapt}

\paragraph{\framework.}
While we implement and evaluate our framework only for the Republic of Korea and the Korean language, \framework generalizes to multiple nation-states in various geo-cultural regions.
National curricula are typically published by governments or educational authorities and are publicly available---\eg United Kingdom\thinspace\footnote{\url{https://www.gov.uk/government/collections/national-curriculum}}, India\thinspace\footnote{\url{http://www.ncert.nic.in/rightside/links/nc_framework.html}}, Bangladesh\thinspace\footnote{\url{https://nctb.gov.bd/}}, and Australia\thinspace\footnote{\url{https://www.australiancurriculum.edu.au/}}.
Nonetheless, we urge practitioners to consider that (1) a nation-state may not follow a single uniform curriculum, with policies and perspectives varying across subgroups; (2) constraining outputs to a specific language may inadvertently shift the framing toward perspectives prevalent regions where that language is widely used~\cite{cho-etal-2025-hermit}; and (3) \framework may inherit and potentially amplify any bias and revisionism latent in the curriculum without a thorough review.

\paragraph{\data.}
We examine the sample-level adaptability of \data to facilitate its extension to other languages and cultures.
Following \citet{jin2024kobbq}, we categorize the queries associated with each learning outcome into three types: (1) \emph{Simply Transferred}, which requires culture-sensitive translation only; (2) \emph{Target Modified}, whose proper nouns and expressions must be modified to the target culture; and (3) \emph{Sample Removed}, which cannot be meaningfully adapted.
Each sample is annotated by three authors who are Korean L1 speakers, resulting in almost perfect inter-annotator agreement (Fleiss’ Kappa~\cite{fleiss1971measuring} = 0.86).
The final labels are determined via majority voting; for samples without a clear majority, all annotators jointly discuss the case to reach a consensus.
Consequently, among 158 learning outcomes, 118 are classified as \emph{Simply Transferred}, 35 as \emph{Target Modified}, and 5 as \emph{Sample Removed}.
All \emph{Sample Removed} learning outcomes are relevant to North Korea, which has a unique political, historical, and geographical relationship to South Korea.
This distribution suggests that most queries and learning outcomes can be adapted to other cultures with minimal effort, highlighting the scalability of \data across languages and cultures.

\subsection{\data Quantitative Analysis}
\label{sec:appendix_quant_analysis}

\paragraph{Culture-specific Query.}

JSD ranges from 0 (identical distributions) to 1 (maximal divergence).
A value of 0.807 in Section~\ref{sec:quant_analysis} indicates strong distributional separation between \data and Korean LIMA in the shared topic space.
To better interpret JSD, we set up two comparison query datasets (\ie KoAlpaca v1.1\thinspace\footnote{\url{https://huggingface.co/datasets/beomi/KoAlpaca-v1.1a}} and \data-filtered) that are culturally agnostic or general and provide qualitative examples for each case.
KoAlpaca v1.1 consists of user-written queries from NAVER Knowledge iN\thinspace\footnote{\url{https://kin.naver.com/}}, a community-driven Q\&A platform in South Korea.
It primarily covers general-domain knowledge acquisition and advising scenarios. 
\data-filtered indicates queries generated from national social studies curricula that are filtered out because they involve general, culture-agnostic knowledge.
We undersample the datasets with a random seed of 42 to match the data size for fair comparison.
We then perform topic modeling and compute JSD compared to Korean LIMA, a standard, well-known general-purpose QA dataset.
While \data and Ko-LIMA yield a JSD of 0.807, implying a substantial topic divergence between the two datasets, KoAlpaca v1.1 and \data-filtered report JSDs of 0.201 and 0.343, respectively, indicating relatively similar topic coverage.
Through qualitative inspection, we observe that Ko-LIMA and KoAlpaca v1.1 is dominated by general-purpose daily life topics (\eg STEM, recipes, and advice), while KoAlpaca v1.1 involves several queries about Korea.
\data-filtered primarily covers general knowledge on social studies (\eg politics, climate, and economy) without any focus on Korean cultural context.
The following example queries showcase each dataset.

\begin{tcolorbox}[breakable, enhanced, top=1pt, left=1pt, right=1pt, bottom=1pt, colback=white, title=Example questions within the filtering process]
\begin{itemize}[leftmargin=*]
    \item \data
    \begin{itemize}[leftmargin=*]
        \item ``\emph{I’m curious about how low birth rates, aging populations, and digitalization are affecting daily life in South Korea these days. How should we adapt to these changes?}''
        \item ``\emph{In South Korea, what do the National Assembly, the executive branch, and the judiciary each do? Why is the separation of powers important for protecting citizens’ rights in South Korea?}''
        \item ``\emph{Can you briefly explain why Dokdo belongs to South Korea, both geographically and historically?}''
    \end{itemize}
    \item Ko-LIMA
    \begin{itemize}[leftmargin=*]
        \item ``\emph{Does Java casting introduce overhead? Or the compiler just resolves everything and there is no cost at run time? Is this a general things, or there are different cases?}''
        \item ``\emph{Write me an official resignation letter.}''
        \item ``\emph{This is my first trip to Shanghai. Please help me plan a classic 3-day tour in Shanghai.}''
    \end{itemize}
    \item KoAlpaca v1.1
    \begin{itemize}[leftmargin=*]
        \item ``\emph{What is the Pygmalion effect? Can you explain it with a specific example?}''
        \item ``\emph{Why is Korea so good at archery? Is there any specific reason?}''
        \item ``\emph{Which bank would you recommend as a primary bank, and why?}''
    \end{itemize}
    \item KCaQA-filtered
    \begin{itemize}[leftmargin=*]
        \item ``\emph{What are the pros and cons of democracy?}''
        \item ``\emph{What can I do as an individual to help tackle climate change and support sustainable development?}''
        \item ``\emph{During inflation, what’s the best place to invest among stocks, bonds, and funds? Could you break down how each one works and what risks they carry?}''
    \end{itemize}
\end{itemize}
\end{tcolorbox}

\paragraph{User-tailored Response.}
Table~\ref{tab:readability} reports readability statistics across languages in terms of response length and lexical diversity, measured by the number of tokens, the number of sentences, tokens per sentence, and the rare token ratio.
\input{sources/table/readability}

\subsection{\data Quality Estimation}
\label{sec:llm_as_a_judge}

\paragraph{LLM-as-a-Judge.}
Table~\ref{tab:llm_as_a_judge} reports the full results of LLM-as-a-Judge for quality estimation of \data.
We further validate these results with a human evaluation.
Specifically, we randomly sample 10 QA instances for each setting and each response level in Korean and English, yielding 120 samples in total.
One of the authors, who is a native Korean speaker and fluent in English, annotates the samples using the same rubric as the LLM-as-a-Judge (Appendix~\ref{sec:rubric}).
We then compute inter-rater reliability between LLM-as-a-Judge and the human annotator.
The results show substantial to almost perfect agreement, with Cohen’s Kappa~\cite{cohen1960coefficient} of 0.98, 0.76, and 0.72 for the three dimensions.
\input{sources/table/llm_as_a_judge}

\paragraph{Multilingual Extension.}

We further validate the translation quality of the multilingual extension of \data queries.
We sample and assess 1k multilingual parallel queries of \data by measuring the COMET-Kiwi~\cite{rei-etal-2022-cometkiwi} score. 
\data achieves COMET-Kiwi scores of 0.941, 0.924, and 0.917 for Ko–En, Ko–Jp, and Ko–Zh, respectively, indicating consistently high-quality translations across languages.
For multilingual responses, we validate semantic similarities among parallel responses.
We sample 1k multilingual parallel responses of \data and compute cross-lingual sentence embeddings using the OpenAI vector embedding model (\ie \texttt{text-embedding-3-large}) and measure semantic similarity across language pairs.
\data yields the semantic similarity scores of 0.911, 0.870, and 0.857 for Ko–En, Ko–Jp, and Ko–Zh, respectively, indicating semantic consistency across languages.

\paragraph{Human Inspection.}
We manually inspect all Korean instances and a randomly sampled set of multilingual instances to qualitatively assess the content and style of \data.
In most cases, queries capture the core topic of the learning objective, and responses align with the corresponding criteria.
However, we observe minor issues in a small fraction of instances, including empty generations, code-switching, and language mismatch.
We also identify cases where the target language appears to shape lexical or cultural choices in the response.
For example, Japanese responses sometimes shift from Korean-standard terminology to expressions preferred in Japan.
While this can be value-neutral when it reflects conventional localization (\eg using `MEXT' instead of `MOE'), we also observe substitutions in disputed, politically contested domains, where responses adopt Japan-aligned terminology (\eg `Takeshima' instead of `Dokdo' or using only `Sea of Japan' while omitting `East Sea').
In addition, some multilingual responses omit details of historically significant events (\eg `March First Movement', a major rebellion against Japanese colonialism).
These observations suggest that the target language may influence response faithfulness and completeness, particularly for culture-specific queries.

\paragraph{Human Validation.}

\input{sources/table/human_validation}

We recruit three Korea-grounded bilingual annotators for each language, rather than target-language native speakers without relevant knowledge of Korean culture.
Specifically, each annotator (1) has been raised in Korea and educated under the Korean national curricula for at least 5 years and (2) demonstrates advanced proficiency and extensive immersion in the target language.
To ensure sufficient linguistic and cultural exposure, we require at least 5 years of residence in the countries where the target language is predominantly spoken (\ie United States, United Kingdom, Canada, Australia, and New Zealand for English; China for Chinese; and Japan for Japanese), as well as official language proficiency test scores indicating an advanced level (\ie TOEFL iBT 110+ for English; HSK Level 6 for Chinese; JLPT N1 for Japanese).
We randomly sample 500 Korean--target-language QA pairs from \data and ask the annotators to assess the data quality.
We adopt the same evaluation setup of LLM-as-a-Judge in Section~\ref{sec:quality_estimation}---\ie binary decision for \emph{language selection} and 1--10 Likert scale scoring for \emph{cultural appropriateness} and \emph{language use}.
Table~\ref{tab:human_validation} reports the human validation results of \data.
The results align with those of LLM-as-a-Judge, yielding almost perfect language selection and more than 8 out of 10 scores for cultural appropriateness and language use, as well as achieving substantial agreement among the three human annotators.

\subsection{Experimental Setting for LLM Training using \data}
\label{sec:apendix_training}

In Section~\ref{sec:training}, we train \texttt{Qwen3-14B} using the following five datasets:
\begin{enumerate}[leftmargin=*]
    \setlength\itemsep{-0.2em}
    \item \textbf{CultureBank:} CultureBank~\cite{shi-etal-2024-culturebank} is a multilingual, multicultural dataset sourced from TikTok and Reddit. We extract the samples related to Korean culture.
    \item \textbf{Korean LIMA:} Korean LIMA is a machine-translated version of the LIMA~\cite{zhou2023lima} dataset, a high-quality, general-purpose open-domain QA.
    \item \textbf{National Curriculum:} This is a source text of \data. Here, we aim to examine the effectiveness of the \framework framework by comparing its results with those of a model trained on \data, a processed output of \framework using the same national curriculum.\thinspace\footnote{This national curriculum can be found at \url{https://ncic.re.kr/}.}
    \item \textbf{\framework without source data:} Instead of using learning outcomes extracted from the national curriculum (Dataset 1), we ask \texttt{GPT-5.2} to generate learning outcomes by itself. Specifically, given non-cultural learning outcomes from other subjects as few-shot examples, we ask \texttt{GPT-5.2} to generate 15 learning outcomes for each of 10 culture topics (Korean history, geography, politics, law, economy, international relations, climate, traditions, social studies, and anthropology). Using these learning outcomes as source data for \framework, we generate queries following steps 3--5 and their responses following steps 1--3.
    \item \textbf{\data (Ours):} We use Korean samples at an advanced level.
\end{enumerate}

Datasets 1, 2, 4, and 5 are open-ended QA tailored for SFT, while Dataset 3 (national curriculum) is a plain corpus.
To match the number of samples used in training, we randomly sample 1,030 instances of Datasets 2–4 with a seed of 42, following the number of the training set of Korean LIMA.
We use accuracy for cultural knowledge evaluation, and follow the A-SVA setting of KorNAT, which asks the models to choose among three options: agree, neutral, and disagree.

\subsection{Replacing Proprietary Models in \framework}
\label{sec:kcaqa_open}

To reduce the reliance on proprietary models, we conduct an additional open-source experiment using \texttt{GLM-5.1} under MIT license as a substitute for the \texttt{GPT-4o} agents used in \framework.
We construct \data-Open by replacing the proprietary generation/evaluation agents with \texttt{GLM-5.1}.
For this small-scale experiment, we simplify the pipeline by omitting the paraphrasing and human-refinement stages, generating only culture-explicit queries, and producing advanced-level responses.
This yields 616 queries across Korean-explicit and three multilingual settings, paired with 2,464 advanced-level responses generated by four open-source LLMs.

\input{sources/table/training_open}

We then randomly sample 1,030 instances from \data-Open and train \texttt{Qwen 3-14B} under the same setup as in Section~\ref{sec:training} and Appendix~\ref{sec:apendix_training}.
Table~\ref{tab:training_open} reports experimental results of LLM post-training (SFT) using \data and \data-Open.
The model trained on \data-Open achieves on-par performance with the original \data and outperforms other baselines on both cultural knowledge and cultural alignment benchmarks.
This suggests that \framework is not tied to proprietary models and can be instantiated with open-source models when accessibility, cost, or reproducibility is prioritized.

We further compute inter-rater reliability between a human annotator and LLM-as-a-Judge using \texttt{GLM 5.1}, following the same setup in Appendix~\ref{sec:llm_as_a_judge} with GPT-Judge.
GLM-Judge also yields substantial to almost perfect agreement, with Cohen’s Kappa of 0.96, 0.78, and 0.70 for three dimensions. 

\subsection{Potential Use Cases}
\label{sec:use_case}

While the primary objective of \framework is to construct culture-specific, open-ended QA datasets for supervised fine-tuning, \framework and \data support several practical scenarios beyond dataset release.

First, we provide SFT-ready supervision for culture-adapted or sovereign LLMs.
This is increasingly important because LLMs often inherit English- or Western-centric priors from training data, leading to cultural misalignment in underrepresented contexts.
\citet{blasi-etal-2022-systematic} shows systematic inequalities across languages and cultures, and \citet{tao2024cultural} reports that LLMs can exhibit cultural values closer to English-speaking or Western populations than to local user communities.
In contrast to translated general instruction data such as Bactrian-X~\cite{li2023bactrianx}, \framework constructs supervision from local institutional sources, making it better suited for culture-grounded post-training.
This use case is empirically supported by our SFT experiment, where training Qwen 3 on \data improves Korean cultural knowledge and alignment over general QA, raw curriculum text, and community-sourced cultural data.

Second, \data can be used to build local civic and educational assistants. Since the dataset is grounded in national social studies curricula, it covers topics such as history, geography, politics, economics, law, social institutions, and international relations, enabling assistants to answer open-ended user questions in ways aligned with local curriculum-based knowledge and civic norms.

Third, \data can serve as a cultural faithfulness evaluation suite: its no-country, implicit, explicit, and multilingual variants allow practitioners to test whether a model preserves the intended local sociocultural context even when the query is phrased indirectly or in another language.

Finally, \framework is a scalable dataset-construction framework for other cultures, as national curricula are often publicly available and provide expert-structured coverage of locally prioritized knowledge. This aligns with recent work on cultural post-training and government/cultural instruction tuning, such as CARE~\cite{guo-etal-2025-care} and Kazakh government/cultural instruction tuning~\cite{laiyk-etal-2025-instruction}.

\section{Detailed Evaluation Rubrics}
\label{sec:rubric}

The evaluation criteria are collaboratively designed by three authors, initially proposed by a single Korean L1 speaker with intermediate English proficiency and elementary proficiency in Chinese and Japanese, and subsequently reviewed and refined by two additional Korean L1 speakers.

\begin{tcolorbox}[breakable, enhanced, top=1pt, left=1pt, right=1pt, bottom=1pt, colback=white, title=Evaluation criteria for both human validation and LLM-as-a-judge]
For the provided query and the generated response, check if they adhere to the following criteria and reject if there is any violence:
\begin{enumerate}[leftmargin=*]
    \item Language selection
    \begin{itemize}[leftmargin=*]
        \item Is the query or the response empty?
        \item Is the query or the response code-mixed with multiple languages, or not written in the target language that is the same with the query?
    \end{itemize}
    \item Cultural appropriateness
    \begin{itemize}[leftmargin=*]
        \item Does the query or the response (especially those are not in Korean) adhere to the desired stance of MOFA (Ministry of Foreign Affairs, Republic of Korea) regarding some sensitive territory issues such as Dokdo (the official terminology that should be used instead of Takeshima) and East Sea (the official terminology that should be written in parallel with Japan Sea)?
        \item Does the query or the response (especially those are not in Korean) include all important details regarding some historical events, \eg the governance and economics during Japanese colonial period?
    \end{itemize}
    \item Language use
    \begin{itemize}[leftmargin=*]
        \item Is the query or the response natural enough to be displayed a colloquial answer for the language speaker?
        \item  For Korean cases, is the synthesized query with/without country-specific expressions (\eg ``our country'') natural enough as a colloquial question?
        \item Does the query or the response contain any terminology that corresponds to a similar concept in Korean, but that should not be transferred to other languages? (\eg the terminology for location, historical events, government departments, etc.)?
    \end{itemize}
\end{enumerate}
\end{tcolorbox}

Throughout the manual inspection, we carefully review the language selection of non-English responses, as LLMs tend to respond in Korean regardless of the input language when the query includes culture-specific content or Korean-specific keywords.
For cultural appropriateness, we refer to official announcements of MOFA (the Ministry of Foreign Affairs, Republic of Korea)\thinspace\footnote{\url{https://www.mofa.go.kr/eng/wpge/m_5441/contents.do}}, for some sensitive issues regarding territory and history (\eg the official terminology for Dokdo and East Sea/Japan Sea).
We further examine whether non-Korean responses include dis/misinformation or blunt any details compared to Korean responses, especially regarding some sensitive, international historical events (\eg the governance and economics during the Japanese colonial period).

\section{System Prompts in \framework}
\label{sec:system_prompt}

\subsection{Query Generation}

\begin{tcolorbox}[breakable, enhanced, top=1pt, left=1pt, right=1pt, bottom=1pt, colback=white, title=Initial query generation (1)]
\begin{Verbatim}[breaklines=true, breaksymbolleft={}, breaksymbolright={}, fontsize=\small]
You are a professional question-generation agent specializing in the Korean K-12 national curriculum.

Task:
  - Convert learning objectives + achievement standards into ONE natural Korean casual question (반말).
  - Must be open-ended, like a real user's curiosity.
  - Avoid formal/academic/LLM-like structure.
  - Use only casual Korean (“~이야?”, “~해줘”, “~알려줘”).

Output exactly ONE Korean casual-speech query.

Learning Objective (Korean): {learning_objectives}
Achievement Standards (Korean): {achievements_standards}

Create one natural Korean user query in casual Korean (반말).
\end{Verbatim}
\end{tcolorbox}

\begin{tcolorbox}[breakable, enhanced, top=1pt, left=1pt, right=1pt, bottom=1pt, colback=white, title=Initial query generation (2)]
\begin{Verbatim}[breaklines=true, breaksymbolleft={}, breaksymbolright={}, fontsize=\small]
You are an expert evaluator for the Korean national curriculum.
  
Evaluate if the query: 
  1) Reflects the learning objective. 
  2) Incorporates key ideas from the achievement standards. 
  3) Sounds natural as Korean casual speech. 
  4) Matches typical Korean user's tone. 
  
Respond ONLY in JSON: 
{"score": 1–5, "suggestion": "brief improvement idea"} 
  
Learning Objective: {learning_objectives} 
Achievement Standards: {achievements_standards} 
Question: {query} 
  
Respond in JSON: 
{"score": 1-5, "suggestion": "brief improvement idea"}
\end{Verbatim}
\end{tcolorbox}

\begin{tcolorbox}[breakable, enhanced, top=1pt, left=1pt, right=1pt, bottom=1pt, colback=white, title=Initial query generation (3)]
\begin{Verbatim}[breaklines=true, breaksymbolleft={}, breaksymbolright={}, fontsize=\small]
Revise the query in natural Korean casual-speech (반말). 
Avoid teacher-like tone and LLM-like structure. 
  
Learning Objective: {learning_objectives} 
Achievement Standards: {achievements_standards} 
Original: {query} 
Feedback: {feedback} 
  
Revise the query.
\end{Verbatim}
\end{tcolorbox}

\begin{tcolorbox}[breakable, enhanced, top=1pt, left=1pt, right=1pt, bottom=1pt, colback=white, title=Culture-sensitive filtering (1)]
\begin{Verbatim}[breaklines=true, breaksymbolleft={}, breaksymbolright={}, fontsize=\small]
You are an expert in Korean culture and shared societal perspectives. 
  
Evaluate whether the question: 
  - Aligns with Korean cultural and social context, 
  - Reflects typical Korean ways of thinking, 
  - Is not a generic universal question, 
  - Is meaningfully grounded in Korean context. 
  
Question: {query} 
  
Respond in JSON: 
{"is_relevant": true/false, "reason": "brief explanation"}
\end{Verbatim}
\end{tcolorbox}

\begin{tcolorbox}[breakable, enhanced, top=1pt, left=1pt, right=1pt, bottom=1pt, colback=white, title=Culture-sensitive filtering (2)]
\begin{Verbatim}[breaklines=true, breaksymbolleft={}, breaksymbolright={}, fontsize=\small]
Polish the query to sound like natural Korean casual-speech (반말). 
  
Question: {query} 
  
Polish this query.
\end{Verbatim}
\end{tcolorbox}

\begin{tcolorbox}[breakable, enhanced, top=1pt, left=1pt, right=1pt, bottom=1pt, colback=white, title=Paraphrasing \& Augmentation]
\begin{Verbatim}[breaklines=true, breaksymbolleft={}, breaksymbolright={}, fontsize=\small]
You are an expert Korean query rewriter. 
Given a base Korean casual-speech query, rewrite it into three Korean variants with different country references. 
  
1) ko_no_country: 
  - No explicit/implicit country mention 
  - Do NOT use "우리나라", "우리 사회", "대한민국", "한국", etc. 
  
2) ko_implicit_country: 
  - No explicit country names 
  - Use implicit expressions like "우리나라", "우리 사회" when appropriate 
  
3) ko_explicit_country: 
  - Explicitly mention "대한민국" or "한국" 
  
All: 
  - Natural Korean casual speech (반말) 
  - Close to original meaning 
  - Respond with strict JSON only. 
  
Base Korean casual query: {query} 
  
Respond in JSON: 
{ "ko_no_country": "...", "ko_implicit_country": "...", "ko_explicit_country": "..." }
\end{Verbatim}
\end{tcolorbox}

\begin{tcolorbox}[breakable, enhanced, top=1pt, left=1pt, right=1pt, bottom=1pt, colback=white, title=Multilingual extension]
\begin{Verbatim}[breaklines=true, breaksymbolleft={}, breaksymbolright={}, fontsize=\small]
Translate the query into English, Simplified Chinese, and Japanese. 
All translations must be natural conversational language. 
  
Respond ONLY in JSON: 
{ "en": "...", "zh": "...", "ja": "..." } 
  
Korean query: {query} 
  
Respond in JSON: 
{ "en": "...", "zh": "...", "ja": "..." } 
\end{Verbatim}
\end{tcolorbox}

\subsection{Response Generation}

\begin{tcolorbox}[breakable, enhanced, top=1pt, left=1pt, right=1pt, bottom=1pt, colback=white, title=User-tailored response generation]
\begin{Verbatim}[breaklines=true, breaksymbolleft={}, breaksymbolright={}, fontsize=\small]
You are a helpful AI assistant specializing in Korean society, culture, history, geography, and civics. 
  
Your purpose is to generate answers that reflect how a person educated under the Korean national curriculum would commonly think and explain social issues: 
  - Empathize before analysis when appropriate, 
  - Consider both personal experience and broader social context, 
  - Balanced and cautious reasoning, 
  - Concrete everyday examples familiar in Korea, 
  - Practical suggestions over abstract theory. 
  
## Role \& Language Rules 
  - The user query can be in Korean, English, Chinese, or Japanese. 
  - EXTREMELY IMPORTANT: You MUST respond strictly and exclusively in the same language as the user query. 
  - Do NOT include any text from other languages (even one word). 
  
## Output Requirement (STRICT JSON) 
Output exactly one valid JSON object with three string fields: 
{{ 
  "basic": "...", 
  "intermediate": "...", 
  "advanced": "..." 
}} 
  
### Basic 
  - Audience: elementary school students and young children 
  - Language: very simple words, short sentences, concrete ideas 
  - Style: everyday examples, one idea per sentence 
  - Content: only the most essential meaning 
  - Length: around 2–3 short sentences 
  
### Intermediate 
  - Audience: high school students and young adults 
  - Language: clear, structured explanations similar to textbooks 
  - Style: balanced reasoning; connect personal experience with broader social context 
  - Content: when relevant, reflect the learning objective and achievement standards (but DO NOT mention them) 
  - Length: around 4–6 sentences 
  
### Advanced 
  - Audience: adults, university/graduate learners, professionals 
  - Language: analytical, evidence-based, conceptually deeper 
  - Style: culturally grounded perspectives common among well-educated Korean adults 
  - Content: contextual analysis, balanced arguments, practical implications; may mention history/institutions/data/research when relevant 
  - Length: around 6–12 sentences, possibly in two short paragraphs  

## Curriculum Reference (DO NOT mention explicitly) 
Learning Objective: {learning_objectives} 
Achievement Standards: {achievements_standards} 
  
Now answer the user query below in the same language as the query.
\end{Verbatim}
\end{tcolorbox}

\begin{tcolorbox}[breakable, enhanced, top=1pt, left=1pt, right=1pt, bottom=1pt, colback=white, title=Response evaluation]
\begin{Verbatim}[breaklines=true, breaksymbolleft={}, breaksymbolright={}, fontsize=\small]
You are a strict evaluator (LLM-as-a-judge). Evaluate the candidate response JSON for the given query. 
  
Return ONLY a valid JSON object with: 
{{ 
  "pass": true/false, 
  "reasons": ["..."], 
  "suggestions": ["..."], 
  "scores": {{ 
    "language_consistency": 1–5, 
    "difficulty_appropriateness": 1–5, 
    "learning_outcome_alignment": 1–5, 
    "cultural_appropriateness": 1–5, 
    "language_use_naturalness": 1–5 
  }} 
}} 
  
Evaluation criteria: 
1) Language selection 
   - Query/response empty? 
   - Code-mixed or not in the same language as the query? 
2) Cultural appropriateness 
   - MOFA-sensitive terminology: 
     * If mentioning Takeshima, must also use Dokdo. 
     * If mentioning Sea of Japan, must include East Sea in parallel. 
   - Ensure major context is not misleading/incorrect for sensitive modern/historical topics. 
3) Language use 
   - Natural colloquial answer for that language. 
   - Avoid transferring Korea-specific terms incorrectly into other languages. 
  
IMPORTANT: 
  - You will also receive "rule_based_findings". 
  - If rule_based_findings is non-empty, you MUST set pass=false and include them in reasons. 
  
Inputs: 
  - Query Language: {query_language} 
  - Learning Objective: {learning_objectives} 
  - Achievement Standards: {achievements_standards} 
  - Query: {query} 
  - Candidate Response JSON: {candidate_json} 
  - Rule-based Findings: {rule_based_findings}
\end{Verbatim}
\end{tcolorbox}

\begin{tcolorbox}[breakable, enhanced, top=1pt, left=1pt, right=1pt, bottom=1pt, colback=white, title=Response revision]
\begin{Verbatim}[breaklines=true, breaksymbolleft={}, breaksymbolright={}, fontsize=\small]
You are revising a response to satisfy strict constraints. 
  
Rules: 
  - Output EXACTLY one JSON object: {{ "basic": "...", "intermediate": "...", "advanced": "..." }} 
  - All values must be strings. 
  - Must be strictly in the same language as the query (no code-mixing).  
  - Must keep Basic/Intermediate/Advanced definitions (length/style/depth). 
  - Must not mention curriculum materials explicitly. 
  - Must fix MOFA-sensitive terminology if relevant: 
    * Use Dokdo (not only Takeshima). 
    * If mentioning Sea of Japan, include East Sea in parallel. 
  
Given: 
  - Query language: {query_language} 
  - Query: {query} 
  - Learning Objective: {learning_objectives} 
  - Achievement Standards: {achievements_standards} 
  - Previous candidate JSON: {candidate_json} 
  - Rule-based findings: {rule_based_findings} 
  - Judge feedback: {judge_json} 
  
Revise the candidate to pass the judge. 
Return JSON only.
\end{Verbatim}
\end{tcolorbox}

%% file: sources/table/readability.tex
\begin{table}[htb!]
\resizebox{\linewidth}{!}{%
\begin{tabular}{@{}lc|rrrr@{}}
\toprule
Lang.               & Lvl.  & \# tokens & \# sentences & TPS  & RTR  \\ \midrule
\multirow{3}{*}{Ko} & B     & 49.6      & 3.3          & 15.5 & 0.14 \\
                    & I     & 113.9     & 4.8          & 23.9 & 0.23 \\
                    & A     & 178.2     & 6.0          & 30.2 & 0.30 \\ \midrule
\multirow{3}{*}{En} & B     & 27.7      & 3.5          & 8.2  & 0.24 \\
                    & I     & 63.6      & 5.0          & 12.8 & 0.25 \\
                    & A     & 103.5     & 6.2          & 17.1 & 0.31 \\ \midrule
\multirow{3}{*}{Zh} & B     & 79.0      & 3.4          & 23.8 & 0.24 \\
                    & I     & 191.5     & 4.8          & 40.4 & 0.26 \\
                    & A     & 313.7     & 6.0          & 53.6 & 0.27 \\ \midrule
\multirow{3}{*}{Ja} & B     & 27.6      & 3.4          & 8.4  & 0.15 \\
                    & I     & 66.5      & 4.9          & 13.8 & 0.21 \\
                    & A     & 112.7     & 6.3          & 18.6 & 0.57 \\ \bottomrule
\end{tabular}%
}
\caption{Readability statistics of \data responses across languages and target levels. B, I, and A indicate basic, intermediate, and advanced levels. TPS and RTR denote token per sentence and rare token ratio, respectively.}
\label{tab:readability}
\end{table}

%% file: sources/table/llm_as_a_judge.tex
\begin{table}[htb!]
\centering
\begin{tabular}{@{}l|l|cccc@{}}
\toprule
Lang.               & Setting                     & Lvl. & LS   & CA   & LU   \\ \midrule
\multirow{9}{*}{Ko} & \multirow{3}{*}{No country} & B    & 1.00 & 9.00 & 7.28 \\
                    &                             & I    & 1.00 & 9.41 & 8.68 \\
                    &                             & A    & 0.98 & 8.98 & 8.28 \\ \cmidrule{2-6}
                    & \multirow{3}{*}{Implicit}   & B    & 1.00 & 8.76 & 7.14 \\
                    &                             & I    & 1.00 & 9.40 & 8.70 \\
                    &                             & A    & 0.99 & 8.93 & 8.29 \\ \cmidrule{2-6}
                    & \multirow{3}{*}{Explicit}   & B    & 1.00 & 8.57 & 6.95 \\
                    &                             & I    & 0.99 & 9.31 & 8.64 \\
                    &                             & A    & 0.98 & 8.93 & 8.28 \\ \midrule
\multirow{3}{*}{En} & \multirow{3}{*}{Explicit}   & B    & 0.99 & 8.00 & 6.64 \\
                    &                             & I    & 0.98 & 8.95 & 8.22 \\
                    &                             & A    & 0.96 & 8.76 & 8.28 \\ \midrule
\multirow{3}{*}{Zh} & \multirow{3}{*}{Explicit}   & B    & 0.81 & 7.78 & 6.66 \\
                    &                             & I    & 0.83 & 8.78 & 8.27 \\
                    &                             & A    & 0.81 & 8.46 & 8.10 \\ \midrule
\multirow{3}{*}{Ja} & \multirow{3}{*}{Explicit}   & B    & 0.94 & 7.86 & 6.88 \\
                    &                             & I    & 0.87 & 8.68 & 8.32 \\
                    &                             & A    & 0.73 & 8.41 & 7.89 \\ \bottomrule
\end{tabular}
\caption{LLM-as-a-judge results for quality estimation of \data. LS, CA, and LU denote the evaluation criteria: Language Selection, Cultural Appropriateness, and Language Use, respectively.}
\label{tab:llm_as_a_judge}
\end{table}

%% file: sources/table/human_validation.tex
\begin{table}[htb!]
\centering
\resizebox{0.7\linewidth}{!}{
\begin{tabular}{@{}l|lccc@{}}
\toprule
Lang.               & Metric   & LS   & CA   & LU   \\ \midrule
\multirow{2}{*}{Ko} & Acc.     & 0.99 & 9.65 & 8.78 \\
                    & $\kappa$ & 0.96 & 0.78 & 0.75 \\ \midrule
\multirow{2}{*}{En} & Acc.     & 1.00 & 8.73 & 8.13 \\
                    & $\kappa$ & 0.98 & 0.71 & 0.66 \\ \midrule
\multirow{2}{*}{Zh} & Acc.     & 0.97 & 9.38 & 8.77 \\
                    & $\kappa$ & 0.99 & 0.84 & 0.72 \\ \midrule
\multirow{2}{*}{Ja} & Acc.     & 0.99 & 9.51 & 8.19 \\
                    & $\kappa$ & 0.96 & 0.79 & 0.68 \\ \bottomrule
\end{tabular}
}
\caption{Human validation results for \data. LS, CA, and LU denote the evaluation criteria: Language Selection, Cultural Appropriateness, and Language Use, respectively. Acc. and $\kappa$ denote the scores and Fleiss' Kappa, respectively.}
\label{tab:human_validation}
\end{table}

%% file: sources/table/training_open.tex
\begin{table}[htb!]
\resizebox{\linewidth}{!}{
\begin{tabular}{@{}lcccc@{}}
\toprule
                            & CLIcK & BLEnD & CulturalBench & KorNAT \\ \midrule
Qwen3-14B                  & 0.587 & 0.842 & 0.585         & 0.679  \\
\hspace{2mm}+ 5. \data      & 0.662 & 0.891 & 0.634         & 0.744  \\
\hspace{2mm}+ 6. \data-Open & 0.658 & 0.887 & 0.631         & 0.742  \\ \bottomrule
\end{tabular}
}
\caption{Experimental results of LLM post-training (SFT) using \data-Open.}
\label{tab:training_open}
\end{table}

%% file: custom.bib
@article{openai2024gpt4,
    title={{GPT}-4 Technical Report}, 
    author={{OpenAI} and Josh Achiam and Steven Adler and Sandhini Agarwal and Lama Ahmad and Ilge Akkaya and Florencia Leoni Aleman and Diogo Almeida and Janko Altenschmidt and Sam Altman and Shyamal Anadkat and Red Avila and Igor Babuschkin and Suchir Balaji and Valerie Balcom and Paul Baltescu and Haiming Bao and Mohammad Bavarian and Jeff Belgum and Irwan Bello and Jake Berdine and Gabriel Bernadett-Shapiro and Christopher Berner and Lenny Bogdonoff and Oleg Boiko and Madelaine Boyd and Anna-Luisa Brakman and Greg Brockman and Tim Brooks and Miles Brundage and Kevin Button and Trevor Cai and Rosie Campbell and Andrew Cann and Brittany Carey and Chelsea Carlson and Rory Carmichael and Brooke Chan and Che Chang and Fotis Chantzis and Derek Chen and Sully Chen and Ruby Chen and Jason Chen and Mark Chen and Ben Chess and Chester Cho and Casey Chu and Hyung Won Chung and Dave Cummings and Jeremiah Currier and Yunxing Dai and Cory Decareaux and Thomas Degry and Noah Deutsch and Damien Deville and Arka Dhar and David Dohan and Steve Dowling and Sheila Dunning and Adrien Ecoffet and Atty Eleti and Tyna Eloundou and David Farhi and Liam Fedus and Niko Felix and Simón Posada Fishman and Juston Forte and Isabella Fulford and Leo Gao and Elie Georges and Christian Gibson and Vik Goel and Tarun Gogineni and Gabriel Goh and Rapha Gontijo-Lopes and Jonathan Gordon and Morgan Grafstein and Scott Gray and Ryan Greene and Joshua Gross and Shixiang Shane Gu and Yufei Guo and Chris Hallacy and Jesse Han and Jeff Harris and Yuchen He and Mike Heaton and Johannes Heidecke and Chris Hesse and Alan Hickey and Wade Hickey and Peter Hoeschele and Brandon Houghton and Kenny Hsu and Shengli Hu and Xin Hu and Joost Huizinga and Shantanu Jain and Shawn Jain and Joanne Jang and Angela Jiang and Roger Jiang and Haozhun Jin and Denny Jin and Shino Jomoto and Billie Jonn and Heewoo Jun and Tomer Kaftan and Łukasz Kaiser and Ali Kamali and Ingmar Kanitscheider and Nitish Shirish Keskar and Tabarak Khan and Logan Kilpatrick and Jong Wook Kim and Christina Kim and Yongjik Kim and Jan Hendrik Kirchner and Jamie Kiros and Matt Knight and Daniel Kokotajlo and Łukasz Kondraciuk and Andrew Kondrich and Aris Konstantinidis and Kyle Kosic and Gretchen Krueger and Vishal Kuo and Michael Lampe and Ikai Lan and Teddy Lee and Jan Leike and Jade Leung and Daniel Levy and Chak Ming Li and Rachel Lim and Molly Lin and Stephanie Lin and Mateusz Litwin and Theresa Lopez and Ryan Lowe and Patricia Lue and Anna Makanju and Kim Malfacini and Sam Manning and Todor Markov and Yaniv Markovski and Bianca Martin and Katie Mayer and Andrew Mayne and Bob McGrew and Scott Mayer McKinney and Christine McLeavey and Paul McMillan and Jake McNeil and David Medina and Aalok Mehta and Jacob Menick and Luke Metz and Andrey Mishchenko and Pamela Mishkin and Vinnie Monaco and Evan Morikawa and Daniel Mossing and Tong Mu and Mira Murati and Oleg Murk and David Mély and Ashvin Nair and Reiichiro Nakano and Rajeev Nayak and Arvind Neelakantan and Richard Ngo and Hyeonwoo Noh and Long Ouyang and Cullen O'Keefe and Jakub Pachocki and Alex Paino and Joe Palermo and Ashley Pantuliano and Giambattista Parascandolo and Joel Parish and Emy Parparita and Alex Passos and Mikhail Pavlov and Andrew Peng and Adam Perelman and Filipe de Avila Belbute Peres and Michael Petrov and Henrique Ponde de Oliveira Pinto and Michael and Pokorny and Michelle Pokrass and Vitchyr H. Pong and Tolly Powell and Alethea Power and Boris Power and Elizabeth Proehl and Raul Puri and Alec Radford and Jack Rae and Aditya Ramesh and Cameron Raymond and Francis Real and Kendra Rimbach and Carl Ross and Bob Rotsted and Henri Roussez and Nick Ryder and Mario Saltarelli and Ted Sanders and Shibani Santurkar and Girish Sastry and Heather Schmidt and David Schnurr and John Schulman and Daniel Selsam and Kyla Sheppard and Toki Sherbakov and Jessica Shieh and Sarah Shoker and Pranav Shyam and Szymon Sidor and Eric Sigler and Maddie Simens and Jordan Sitkin and Katarina Slama and Ian Sohl and Benjamin Sokolowsky and Yang Song and Natalie Staudacher and Felipe Petroski Such and Natalie Summers and Ilya Sutskever and Jie Tang and Nikolas Tezak and Madeleine B. Thompson and Phil Tillet and Amin Tootoonchian and Elizabeth Tseng and Preston Tuggle and Nick Turley and Jerry Tworek and Juan Felipe Cerón Uribe and Andrea Vallone and Arun Vijayvergiya and Chelsea Voss and Carroll Wainwright and Justin Jay Wang and Alvin Wang and Ben Wang and Jonathan Ward and Jason Wei and CJ Weinmann and Akila Welihinda and Peter Welinder and Jiayi Weng and Lilian Weng and Matt Wiethoff and Dave Willner and Clemens Winter and Samuel Wolrich and Hannah Wong and Lauren Workman and Sherwin Wu and Jeff Wu and Michael Wu and Kai Xiao and Tao Xu and Sarah Yoo and Kevin Yu and Qiming Yuan and Wojciech Zaremba and Rowan Zellers and Chong Zhang and Marvin Zhang and Shengjia Zhao and Tianhao Zheng and Juntang Zhuang and William Zhuk and Barret Zoph},
    year={2024},
    journal={arXiv preprint arXiv:2303.08774},
    url={https://arxiv.org/abs/2303.08774}, 
}

@article{openai2025gptoss,
    title={gpt-oss-120b \& gpt-oss-20b Model Card}, 
    author={{OpenAI} and Sandhini Agarwal and Lama Ahmad and Jason Ai and Sam Altman and Andy Applebaum and Edwin Arbus and Rahul K. Arora and Yu Bai and Bowen Baker and Haiming Bao and Boaz Barak and Ally Bennett and Tyler Bertao and Nivedita Brett and Eugene Brevdo and Greg Brockman and Sebastien Bubeck and Che Chang and Kai Chen and Mark Chen and Enoch Cheung and Aidan Clark and Dan Cook and Marat Dukhan and Casey Dvorak and Kevin Fives and Vlad Fomenko and Timur Garipov and Kristian Georgiev and Mia Glaese and Tarun Gogineni and Adam Goucher and Lukas Gross and Katia Gil Guzman and John Hallman and Jackie Hehir and Johannes Heidecke and Alec Helyar and Haitang Hu and Romain Huet and Jacob Huh and Saachi Jain and Zach Johnson and Chris Koch and Irina Kofman and Dominik Kundel and Jason Kwon and Volodymyr Kyrylov and Elaine Ya Le and Guillaume Leclerc and James Park Lennon and Scott Lessans and Mario Lezcano-Casado and Yuanzhi Li and Zhuohan Li and Ji Lin and Jordan Liss and Lily and Liu and Jiancheng Liu and Kevin Lu and Chris Lu and Zoran Martinovic and Lindsay McCallum and Josh McGrath and Scott McKinney and Aidan McLaughlin and Song Mei and Steve Mostovoy and Tong Mu and Gideon Myles and Alexander Neitz and Alex Nichol and Jakub Pachocki and Alex Paino and Dana Palmie and Ashley Pantuliano and Giambattista Parascandolo and Jongsoo Park and Leher Pathak and Carolina Paz and Ludovic Peran and Dmitry Pimenov and Michelle Pokrass and Elizabeth Proehl and Huida Qiu and Gaby Raila and Filippo Raso and Hongyu Ren and Kimmy Richardson and David Robinson and Bob Rotsted and Hadi Salman and Suvansh Sanjeev and Max Schwarzer and D. Sculley and Harshit Sikchi and Kendal Simon and Karan Singhal and Yang Song and Dane Stuckey and Zhiqing Sun and Philippe Tillet and Sam Toizer and Foivos Tsimpourlas and Nikhil Vyas and Eric Wallace and Xin Wang and Miles Wang and Olivia Watkins and Kevin Weil and Amy Wendling and Kevin Whinnery and Cedric Whitney and Hannah Wong and Lin Yang and Yu Yang and Michihiro Yasunaga and Kristen Ying and Wojciech Zaremba and Wenting Zhan and Cyril Zhang and Brian Zhang and Eddie Zhang and Shengjia Zhao},
    year={2025},
    journal={arXiv preprint arXiv:2508.10925},
    url={https://arxiv.org/abs/2508.10925}, 
}

@article{yang2025qwen3,
    title={Qwen3 Technical Report}, 
    author={An Yang and Anfeng Li and Baosong Yang and Beichen Zhang and Binyuan Hui and Bo Zheng and Bowen Yu and Chang Gao and Chengen Huang and Chenxu Lv and Chujie Zheng and Dayiheng Liu and Fan Zhou and Fei Huang and Feng Hu and Hao Ge and Haoran Wei and Huan Lin and Jialong Tang and Jian Yang and Jianhong Tu and Jianwei Zhang and Jianxin Yang and Jiaxi Yang and Jing Zhou and Jingren Zhou and Junyang Lin and Kai Dang and Keqin Bao and Kexin Yang and Le Yu and Lianghao Deng and Mei Li and Mingfeng Xue and Mingze Li and Pei Zhang and Peng Wang and Qin Zhu and Rui Men and Ruize Gao and Shixuan Liu and Shuang Luo and Tianhao Li and Tianyi Tang and Wenbiao Yin and Xingzhang Ren and Xinyu Wang and Xinyu Zhang and Xuancheng Ren and Yang Fan and Yang Su and Yichang Zhang and Yinger Zhang and Yu Wan and Yuqiong Liu and Zekun Wang and Zeyu Cui and Zhenru Zhang and Zhipeng Zhou and Zihan Qiu},
    year={2025},
    journal={arXiv preprint arXiv:2505.09388},
    url={https://arxiv.org/abs/2505.09388}, 
}

@article{deepseekai2025deepseekr1,
    title={{D}eep{S}eek-{R}1: Incentivizing Reasoning Capability in {LLM}s via Reinforcement Learning}, 
    author={{DeepSeek-AI} and Daya Guo and Dejian Yang and Haowei Zhang and Junxiao Song and Ruoyu Zhang and Runxin Xu and Qihao Zhu and Shirong Ma and Peiyi Wang and Xiao Bi and Xiaokang Zhang and Xingkai Yu and Yu Wu and Z. F. Wu and Zhibin Gou and Zhihong Shao and Zhuoshu Li and Ziyi Gao and Aixin Liu and Bing Xue and Bingxuan Wang and Bochao Wu and Bei Feng and Chengda Lu and Chenggang Zhao and Chengqi Deng and Chenyu Zhang and Chong Ruan and Damai Dai and Deli Chen and Dongjie Ji and Erhang Li and Fangyun Lin and Fucong Dai and Fuli Luo and Guangbo Hao and Guanting Chen and Guowei Li and H. Zhang and Han Bao and Hanwei Xu and Haocheng Wang and Honghui Ding and Huajian Xin and Huazuo Gao and Hui Qu and Hui Li and Jianzhong Guo and Jiashi Li and Jiawei Wang and Jingchang Chen and Jingyang Yuan and Junjie Qiu and Junlong Li and J. L. Cai and Jiaqi Ni and Jian Liang and Jin Chen and Kai Dong and Kai Hu and Kaige Gao and Kang Guan and Kexin Huang and Kuai Yu and Lean Wang and Lecong Zhang and Liang Zhao and Litong Wang and Liyue Zhang and Lei Xu and Leyi Xia and Mingchuan Zhang and Minghua Zhang and Minghui Tang and Meng Li and Miaojun Wang and Mingming Li and Ning Tian and Panpan Huang and Peng Zhang and Qiancheng Wang and Qinyu Chen and Qiushi Du and Ruiqi Ge and Ruisong Zhang and Ruizhe Pan and Runji Wang and R. J. Chen and R. L. Jin and Ruyi Chen and Shanghao Lu and Shangyan Zhou and Shanhuang Chen and Shengfeng Ye and Shiyu Wang and Shuiping Yu and Shunfeng Zhou and Shuting Pan and S. S. Li and Shuang Zhou and Shaoqing Wu and Shengfeng Ye and Tao Yun and Tian Pei and Tianyu Sun and T. Wang and Wangding Zeng and Wanjia Zhao and Wen Liu and Wenfeng Liang and Wenjun Gao and Wenqin Yu and Wentao Zhang and W. L. Xiao and Wei An and Xiaodong Liu and Xiaohan Wang and Xiaokang Chen and Xiaotao Nie and Xin Cheng and Xin Liu and Xin Xie and Xingchao Liu and Xinyu Yang and Xinyuan Li and Xuecheng Su and Xuheng Lin and X. Q. Li and Xiangyue Jin and Xiaojin Shen and Xiaosha Chen and Xiaowen Sun and Xiaoxiang Wang and Xinnan Song and Xinyi Zhou and Xianzu Wang and Xinxia Shan and Y. K. Li and Y. Q. Wang and Y. X. Wei and Yang Zhang and Yanhong Xu and Yao Li and Yao Zhao and Yaofeng Sun and Yaohui Wang and Yi Yu and Yichao Zhang and Yifan Shi and Yiliang Xiong and Ying He and Yishi Piao and Yisong Wang and Yixuan Tan and Yiyang Ma and Yiyuan Liu and Yongqiang Guo and Yuan Ou and Yuduan Wang and Yue Gong and Yuheng Zou and Yujia He and Yunfan Xiong and Yuxiang Luo and Yuxiang You and Yuxuan Liu and Yuyang Zhou and Y. X. Zhu and Yanhong Xu and Yanping Huang and Yaohui Li and Yi Zheng and Yuchen Zhu and Yunxian Ma and Ying Tang and Yukun Zha and Yuting Yan and Z. Z. Ren and Zehui Ren and Zhangli Sha and Zhe Fu and Zhean Xu and Zhenda Xie and Zhengyan Zhang and Zhewen Hao and Zhicheng Ma and Zhigang Yan and Zhiyu Wu and Zihui Gu and Zijia Zhu and Zijun Liu and Zilin Li and Ziwei Xie and Ziyang Song and Zizheng Pan and Zhen Huang and Zhipeng Xu and Zhongyu Zhang and Zhen Zhang},
    year={2025},
    journal={arXiv preprint arXiv:2501.12948},
    url={https://arxiv.org/abs/2501.12948}, 
}

@article{jin2024kobbq,
    author = {Jin, Jiho and Kim, Jiseon and Lee, Nayeon and Yoo, Haneul and Oh, Alice and Lee, Hwaran},
    title = {Ko{BBQ}: Korean Bias Benchmark for Question Answering},
    journal = {Transactions of the Association for Computational Linguistics},
    volume = {12},
    pages = {507-524},
    year = {2024},
    month = {05},
    issn = {2307-387X},
    doi = {10.1162/tacl_a_00661},
    url = {https://doi.org/10.1162/tacl_a_00661},
    eprint = {https://direct.mit.edu/tacl/article-pdf/doi/10.1162/tacl_a_00661/2369542/tacl_a_00661.pdf},
}

@article{gunasekar2023textbooks,
    title={Textbooks are all you need},
    author={Suriya Gunasekar and Yi Zhang and Jyoti Aneja and Caio César Teodoro Mendes and Allie Del Giorno and Sivakanth Gopi and Mojan Javaheripi and Piero Kauffmann and Gustavo de Rosa and Olli Saarikivi and Adil Salim and Shital Shah and Harkirat Singh Behl and Xin Wang and Sébastien Bubeck and Ronen Eldan and Adam Tauman Kalai and Yin Tat Lee and Yuanzhi Li},
    journal={arXiv preprint arXiv:2306.11644},
    url={https://arxiv.org/abs/2306.11644}, 
    year={2023}
}

@misc{benallal2024cosmopedia,
    author = {Ben Allal, Loubna and Lozhkov, Anton and Penedo, Guilherme and Wolf, Thomas and von Werra, Leandro},
    title = {Cosmopedia},
    month = {February},
    year = {2024},
    url = {https://huggingface.co/datasets/HuggingFaceTB/cosmopedia}
}

@inproceedings{wang2024codeclm,
    title = "{C}odec{LM}: Aligning Language Models with Tailored Synthetic Data",
    author = "Wang, Zifeng  and
      Li, Chun-Liang  and
      Perot, Vincent  and
      Le, Long  and
      Miao, Jin  and
      Zhang, Zizhao  and
      Lee, Chen-Yu  and
      Pfister, Tomas",
    editor = "Duh, Kevin  and
      Gomez, Helena  and
      Bethard, Steven",
    booktitle = "Findings of the Association for Computational Linguistics: NAACL 2024",
    month = jun,
    year = "2024",
    address = "Mexico City, Mexico",
    publisher = "Association for Computational Linguistics",
    url = "https://aclanthology.org/2024.findings-naacl.235/",
    doi = "10.18653/v1/2024.findings-naacl.235",
    pages = "3712--3729",
}

@inproceedings{jones2024teaching,
    title={Teaching Language Models to Hallucinate Less with Synthetic Tasks},
    author={Erik Jones and Hamid Palangi and Clarisse Sim{\~o}es Ribeiro and Varun Chandrasekaran and Subhabrata Mukherjee and Arindam Mitra and Ahmed Hassan Awadallah and Ece Kamar},
    booktitle={The Twelfth International Conference on Learning Representations},
    year={2024},
    url={https://openreview.net/forum?id=xpw7V0P136}
}

@article{maini2025beyondweb,
    title={Beyond{W}eb: Lessons from scaling synthetic data for trillion-scale pretraining},
    author={Maini, Pratyush and Dorna, Vineeth and Doshi, Parth and Carranza, Aldo and Pan, Fan and Urbanek, Jack and Burstein, Paul and Fang, Alex and Deng, Alvin and Abbas, Amro and others},
    journal={arXiv preprint arXiv:2508.10975},
    url={https://arxiv.org/abs/2508.10975}, 
    year={2025}
}

@article{spencer2012culture,
    title={What is culture},
    author={Spencer-Oatey, Helen and Franklin, Peter},
    journal={A compilation of quotations. GlobalPAD Core Concepts},
    volume={1},
    number={22},
    pages={1--21},
    year={2012}
}

@inproceedings{rao2025normad,
    title = "{N}orm{A}d: A Framework for Measuring the Cultural Adaptability of Large Language Models",
    author = "Rao, Abhinav Sukumar  and
      Yerukola, Akhila  and
      Shah, Vishwa  and
      Reinecke, Katharina  and
      Sap, Maarten",
    editor = "Chiruzzo, Luis  and
      Ritter, Alan  and
      Wang, Lu",
    booktitle = "Proceedings of the 2025 Conference of the Nations of the Americas Chapter of the Association for Computational Linguistics: Human Language Technologies (Volume 1: Long Papers)",
    month = apr,
    year = "2025",
    address = "Albuquerque, New Mexico",
    publisher = "Association for Computational Linguistics",
    url = "https://aclanthology.org/2025.naacl-long.120/",
    doi = "10.18653/v1/2025.naacl-long.120",
    pages = "2373--2403",
    ISBN = "979-8-89176-189-6",
}

@inproceedings{li2024culturepark,
    author = {Li, Cheng and Teney, Damien and Yang, Linyi and Wen, Qingsong and Xie, Xing and Wang, Jindong},
    booktitle = {Advances in Neural Information Processing Systems},
    doi = {10.52202/079017-2082},
    editor = {A. Globerson and L. Mackey and D. Belgrave and A. Fan and U. Paquet and J. Tomczak and C. Zhang},
    pages = {65183--65216},
    publisher = {Curran Associates, Inc.},
    title = {Culture{P}ark: Boosting Cross-cultural Understanding in Large Language Models},
    url = {https://proceedings.neurips.cc/paper_files/paper/2024/file/77f089cd16dbc36ddd1caeb18446fbdd-Paper-Conference.pdf},
    volume = {37},
    year = {2024}
}

@inproceedings{tanwar2025you,
    title = "Do You Know About My Nation? Investigating Multilingual Language Models' Cultural Literacy Through Factual Knowledge",
    author = "Tanwar, Eshaan  and
      Chatterjee, Anwoy  and
      Saxon, Michael  and
      Albalak, Alon  and
      Wang, William Yang  and
      Chakraborty, Tanmoy",
    editor = "Christodoulopoulos, Christos  and
      Chakraborty, Tanmoy  and
      Rose, Carolyn  and
      Peng, Violet",
    booktitle = "Proceedings of the 2025 Conference on Empirical Methods in Natural Language Processing",
    month = nov,
    year = "2025",
    address = "Suzhou, China",
    publisher = "Association for Computational Linguistics",
    url = "https://aclanthology.org/2025.emnlp-main.756/",
    doi = "10.18653/v1/2025.emnlp-main.756",
    pages = "14967--14990",
    ISBN = "979-8-89176-332-6",
}

@inproceedings{lee2024kornat,
    title = "{K}or{NAT}: {LLM} Alignment Benchmark for {K}orean Social Values and Common Knowledge",
    author = "Lee, Jiyoung  and
      Kim, Minwoo  and
      Kim, Seungho  and
      Kim, Junghwan  and
      Won, Seunghyun  and
      Lee, Hwaran  and
      Choi, Edward",
    editor = "Ku, Lun-Wei  and
      Martins, Andre  and
      Srikumar, Vivek",
    booktitle = "Findings of the Association for Computational Linguistics: ACL 2024",
    month = aug,
    year = "2024",
    address = "Bangkok, Thailand",
    publisher = "Association for Computational Linguistics",
    url = "https://aclanthology.org/2024.findings-acl.666/",
    doi = "10.18653/v1/2024.findings-acl.666",
    pages = "11177--11213",
}

@inproceedings{kim2022kochet,
    title = "{K}o{CHET}: A {K}orean Cultural Heritage Corpus for Entity-related Tasks",
    author = "Kim, Gyeongmin  and
      Kim, Jinsung  and
      Son, Junyoung  and
      Lim, Heuiseok",
    editor = "Calzolari, Nicoletta  and
      Huang, Chu-Ren  and
      Kim, Hansaem  and
      Pustejovsky, James  and
      Wanner, Leo  and
      Choi, Key-Sun  and
      Ryu, Pum-Mo  and
      Chen, Hsin-Hsi  and
      Donatelli, Lucia  and
      Ji, Heng  and
      Kurohashi, Sadao  and
      Paggio, Patrizia  and
      Xue, Nianwen  and
      Kim, Seokhwan  and
      Hahm, Younggyun  and
      He, Zhong  and
      Lee, Tony Kyungil  and
      Santus, Enrico  and
      Bond, Francis  and
      Na, Seung-Hoon",
    booktitle = "Proceedings of the 29th International Conference on Computational Linguistics",
    month = oct,
    year = "2022",
    address = "Gyeongju, Republic of Korea",
    publisher = "International Committee on Computational Linguistics",
    url = "https://aclanthology.org/2022.coling-1.308/",
    pages = "3496--3505",
}

@inproceedings{son2024hae,
    title = "{HAE}-{RAE} Bench: Evaluation of {K}orean Knowledge in Language Models",
    author = "Son, Guijin  and
      Lee, Hanwool  and
      Kim, Suwan  and
      Kim, Huiseo  and
      Lee, Jae cheol  and
      Yeom, Je Won  and
      Jung, Jihyu  and
      Kim, Jung woo  and
      Kim, Songseong",
    editor = "Calzolari, Nicoletta  and
      Kan, Min-Yen  and
      Hoste, Veronique  and
      Lenci, Alessandro  and
      Sakti, Sakriani  and
      Xue, Nianwen",
    booktitle = "Proceedings of the 2024 Joint International Conference on Computational Linguistics, Language Resources and Evaluation (LREC-COLING 2024)",
    month = may,
    year = "2024",
    address = "Torino, Italia",
    publisher = "ELRA and ICCL",
    url = "https://aclanthology.org/2024.lrec-main.704/",
    pages = "7993--8007",
}

@article{ko2025kosimpleqa,
    title={Ko{S}imple{QA}: A {K}orean Factuality Benchmark with an Analysis of Reasoning {LLM}s},
    author={Ko, Donghyeon and Jin, Yeguk and Chae, Kyubyung and Lee, Byungwook and Jo, Chansong and In, Sookyo and Lee, Jaehong and Kim, Taesup and Kwak, Donghyun},
    journal={arXiv preprint arXiv:2510.18368},
    url={https://arxiv.org/abs/2510.18368}, 
    year={2025}
}

@inproceedings{zhent2023judging,
    author = {Zheng, Lianmin and Chiang, Wei-Lin and Sheng, Ying and Zhuang, Siyuan and Wu, Zhanghao and Zhuang, Yonghao and Lin, Zi and Li, Zhuohan and Li, Dacheng and Xing, Eric and Zhang, Hao and Gonzalez, Joseph E and Stoica, Ion},
    booktitle = {Advances in Neural Information Processing Systems},
    editor = {A. Oh and T. Naumann and A. Globerson and K. Saenko and M. Hardt and S. Levine},
    pages = {46595--46623},
    publisher = {Curran Associates, Inc.},
    title = {Judging {LLM}-as-a-Judge with {MT}-Bench and Chatbot Arena},
    url = {https://proceedings.neurips.cc/paper_files/paper/2023/file/91f18a1287b398d378ef22505bf41832-Paper-Datasets_and_Benchmarks.pdf},
    volume = {36},
    year = {2023}
}

@misc{robyn2022wordfreq,
    author       = {Robyn Speer},
    title        = {rspeer/wordfreq: v3.0},
    month        = sep,
    year         = 2022,
    publisher    = {Zenodo},
    version      = {v3.0.2},
    doi          = {10.5281/zenodo.7199437},
    url          = {https://doi.org/10.5281/zenodo.7199437}
}

@inproceedings{zhou2023lima,
    author = {Zhou, Chunting and Liu, Pengfei and Xu, Puxin and Iyer, Srinivasan and Sun, Jiao and Mao, Yuning and Ma, Xuezhe and Efrat, Avia and Yu, Ping and YU, LILI and Zhang, Susan and Ghosh, Gargi and Lewis, Mike and Zettlemoyer, Luke and Levy, Omer},
    booktitle = {Advances in Neural Information Processing Systems},
    editor = {A. Oh and T. Naumann and A. Globerson and K. Saenko and M. Hardt and S. Levine},
    pages = {55006--55021},
    publisher = {Curran Associates, Inc.},
    title = {LIMA: Less Is More for Alignment},
    url = {https://proceedings.neurips.cc/paper_files/paper/2023/file/ac662d74829e4407ce1d126477f4a03a-Paper-Conference.pdf},
    volume = {36},
    year = {2023}
}

@misc{upstage2025solarpro2,
    title={Solar {P}ro 2},
    author={Upstage},
    year={2025},
    url={https://www.upstage.ai/blog/en/solar-pro-2-launch}
}

@misc{kr-sbert,
    author = {Park, Suzi and Hyopil Shin},
    title = {KR-SBERT: A Pre-trained Korean-specific Sentence-BERT model},
    year = {2021},
    publisher = {GitHub},
    journal = {GitHub repository},
    howpublished = {\url{https://github.com/snunlp/KR-SBERT}}
}

@article{grootendorst2022bertopic,
    title={{BERT}opic: Neural topic modeling with a class-based {TF-IDF} procedure},
    author={Grootendorst, Maarten},
    journal={arXiv preprint arXiv:2203.05794},
    year={2022},
    url={https://arxiv.org/abs/2203.05794}
}

@inproceedings{lee-etal-2023-kosbi,
    title = "{K}o{SBI}: A Dataset for Mitigating Social Bias Risks Towards Safer Large Language Model Applications",
    author = "Lee, Hwaran  and
      Hong, Seokhee  and
      Park, Joonsuk  and
      Kim, Takyoung  and
      Kim, Gunhee  and
      Ha, Jung-woo",
    editor = "Sitaram, Sunayana  and
      Beigman Klebanov, Beata  and
      Williams, Jason D",
    booktitle = "Proceedings of the 61st Annual Meeting of the Association for Computational Linguistics (Volume 5: Industry Track)",
    month = jul,
    year = "2023",
    address = "Toronto, Canada",
    publisher = "Association for Computational Linguistics",
    url = "https://aclanthology.org/2023.acl-industry.21/",
    doi = "10.18653/v1/2023.acl-industry.21",
    pages = "208--224",
}

@inproceedings{blasi-etal-2022-systematic,
    title = "Systematic Inequalities in Language Technology Performance across the World{'}s Languages",
    author = "Blasi, Damian  and
      Anastasopoulos, Antonios  and
      Neubig, Graham",
    editor = "Muresan, Smaranda  and
      Nakov, Preslav  and
      Villavicencio, Aline",
    booktitle = "Proceedings of the 60th Annual Meeting of the Association for Computational Linguistics (Volume 1: Long Papers)",
    month = may,
    year = "2022",
    address = "Dublin, Ireland",
    publisher = "Association for Computational Linguistics",
    url = "https://aclanthology.org/2022.acl-long.376/",
    doi = "10.18653/v1/2022.acl-long.376",
    pages = "5486--5505",
}

@article{qin2025survey,
    title = {A survey of multilingual large language models},
    journal = {Patterns},
    volume = {6},
    number = {1},
    pages = {101118},
    year = {2025},
    issn = {2666-3899},
    doi = {https://doi.org/10.1016/j.patter.2024.101118},
    url = {https://www.sciencedirect.com/science/article/pii/S2666389924002903},
    author = {Libo Qin and Qiguang Chen and Yuhang Zhou and Zhi Chen and Yinghui Li and Lizi Liao and Min Li and Wanxiang Che and Philip S. Yu},
}

@article{zhu2024multilingual,
    title={Multilingual Large Language Models: A Systematic Survey}, 
    author={Shaolin Zhu and Supryadi and Shaoyang Xu and Haoran Sun and Leiyu Pan and Menglong Cui and Jiangcun Du and Renren Jin and António Branco and Deyi Xiong},
    year={2024},
    journal={arXiv preprint arXiv:2411.11072},
    url={https://arxiv.org/abs/2411.11072}, 
}

@article{held2023material,
    title={A Material Lens on Coloniality in {NLP}}, 
    author={William Held and Camille Harris and Michael Best and Diyi Yang},
    year={2023},
    journal={arXiv preprint arXiv:2311.08391},
    url={https://arxiv.org/abs/2311.08391}, 
}

@inproceedings{alkhamissi-etal-2024-investigating,
    title = "Investigating Cultural Alignment of Large Language Models",
    author = "AlKhamissi, Badr  and
      ElNokrashy, Muhammad  and
      Alkhamissi, Mai  and
      Diab, Mona",
    editor = "Ku, Lun-Wei  and
      Martins, Andre  and
      Srikumar, Vivek",
    booktitle = "Proceedings of the 62nd Annual Meeting of the Association for Computational Linguistics (Volume 1: Long Papers)",
    month = aug,
    year = "2024",
    address = "Bangkok, Thailand",
    publisher = "Association for Computational Linguistics",
    url = "https://aclanthology.org/2024.acl-long.671/",
    doi = "10.18653/v1/2024.acl-long.671",
    pages = "12404--12422",
}

@article{tao2024cultural,
    author = {Tao, Yan and Viberg, Olga and Baker, Ryan S and Kizilcec, René F},
    title = {Cultural bias and cultural alignment of large language models},
    journal = {PNAS Nexus},
    volume = {3},
    number = {9},
    pages = {pgae346},
    year = {2024},
    month = {09},
    issn = {2752-6542},
    doi = {10.1093/pnasnexus/pgae346},
    url = {https://doi.org/10.1093/pnasnexus/pgae346},
    eprint = {https://academic.oup.com/pnasnexus/article-pdf/3/9/pgae346/59151559/pgae346.pdf},
}

@inproceedings{xu-etal-2025-self,
    title = "Self-Pluralising Culture Alignment for Large Language Models",
    author = "Xu, Shaoyang  and
      Leng, Yongqi  and
      Yu, Linhao  and
      Xiong, Deyi",
    editor = "Chiruzzo, Luis  and
      Ritter, Alan  and
      Wang, Lu",
    booktitle = "Proceedings of the 2025 Conference of the Nations of the Americas Chapter of the Association for Computational Linguistics: Human Language Technologies (Volume 1: Long Papers)",
    month = apr,
    year = "2025",
    address = "Albuquerque, New Mexico",
    publisher = "Association for Computational Linguistics",
    url = "https://aclanthology.org/2025.naacl-long.350/",
    doi = "10.18653/v1/2025.naacl-long.350",
    pages = "6859--6877",
    ISBN = "979-8-89176-189-6",
}

@inproceedings{masoud-etal-2025-cultural,
    title = "Cultural Alignment in Large Language Models: An Explanatory Analysis Based on Hofstede{'}s Cultural Dimensions",
    author = "Masoud, Reem  and
      Liu, Ziquan  and
      Ferianc, Martin  and
      Treleaven, Philip C.  and
      Rodrigues, Miguel Rodrigues",
    editor = "Rambow, Owen  and
      Wanner, Leo  and
      Apidianaki, Marianna  and
      Al-Khalifa, Hend  and
      Eugenio, Barbara Di  and
      Schockaert, Steven",
    booktitle = "Proceedings of the 31st International Conference on Computational Linguistics",
    month = jan,
    year = "2025",
    address = "Abu Dhabi, UAE",
    publisher = "Association for Computational Linguistics",
    url = "https://aclanthology.org/2025.coling-main.567/",
    pages = "8474--8503",
}

@inproceedings{romanou2025include,
    title={{INCLUDE}: Evaluating Multilingual Language Understanding with Regional Knowledge},
    author={Angelika Romanou and Negar Foroutan and Anna Sotnikova and Sree Harsha Nelaturu and Shivalika Singh and Rishabh Maheshwary and Micol Altomare and Zeming Chen and Mohamed A. Haggag and Snegha A and Alfonso Amayuelas and Azril Hafizi Amirudin and Danylo Boiko and Michael Chang and Jenny Chim and Gal Cohen and Aditya Kumar Dalmia and Abraham Diress and Sharad Duwal and Daniil Dzenhaliou and Daniel Fernando Erazo Florez and Fabian Farestam and Joseph Marvin Imperial and Shayekh Bin Islam and Perttu Isotalo and Maral Jabbarishiviari and B{\"o}rje F. Karlsson and Eldar Khalilov and Christopher Klamm and Fajri Koto and Dominik Krzemi{\'n}ski and Gabriel Adriano de Melo and Syrielle Montariol and Yiyang Nan and Joel Niklaus and Jekaterina Novikova and Johan Samir Obando Ceron and Debjit Paul and Esther Ploeger and Jebish Purbey and Swati Rajwal and Selvan Sunitha Ravi and Sara Rydell and Roshan Santhosh and Drishti Sharma and Marjana Prifti Skenduli and Arshia Soltani Moakhar and Bardia soltani moakhar and Ayush Kumar Tarun and Azmine Toushik Wasi and Thenuka Ovin Weerasinghe and Serhan Yilmaz and Mike Zhang and Imanol Schlag and Marzieh Fadaee and Sara Hooker and Antoine Bosselut},
    booktitle={The Thirteenth International Conference on Learning Representations},
    year={2025},
    url={https://openreview.net/forum?id=k3gCieTXeY}
}

@inproceedings{kwok2024evaluating,
    title={Evaluating Cultural Adaptability of a Large Language Model via Simulation of Synthetic Personas},
    author={Louis Kwok and Michal Bravansky and Lewis Griffin},
    booktitle={First Conference on Language Modeling},
    year={2024},
    url={https://openreview.net/forum?id=S4ZOkV1AHl}
}

@inproceedings{liu-etal-2024-multilingual,
    title = "Are Multilingual {LLM}s Culturally-Diverse Reasoners? An Investigation into Multicultural Proverbs and Sayings",
    author = "Cecilia Liu, Chen  and
      Koto, Fajri  and
      Baldwin, Timothy  and
      Gurevych, Iryna",
    editor = "Duh, Kevin  and
      Gomez, Helena  and
      Bethard, Steven",
    booktitle = "Proceedings of the 2024 Conference of the North American Chapter of the Association for Computational Linguistics: Human Language Technologies (Volume 1: Long Papers)",
    month = jun,
    year = "2024",
    address = "Mexico City, Mexico",
    publisher = "Association for Computational Linguistics",
    url = "https://aclanthology.org/2024.naacl-long.112/",
    doi = "10.18653/v1/2024.naacl-long.112",
    pages = "2016--2039",
}

@inproceedings{foroutan-etal-2022-discovering,
    title = "Discovering Language-neutral Sub-networks in Multilingual Language Models",
    author = "Foroutan, Negar  and
      Banaei, Mohammadreza  and
      Lebret, R{\'e}mi  and
      Bosselut, Antoine  and
      Aberer, Karl",
    editor = "Goldberg, Yoav  and
      Kozareva, Zornitsa  and
      Zhang, Yue",
    booktitle = "Proceedings of the 2022 Conference on Empirical Methods in Natural Language Processing",
    month = dec,
    year = "2022",
    address = "Abu Dhabi, United Arab Emirates",
    publisher = "Association for Computational Linguistics",
    url = "https://aclanthology.org/2022.emnlp-main.513/",
    doi = "10.18653/v1/2022.emnlp-main.513",
    pages = "7560--7575",
}

@article{fleiss1971measuring,
    author={Fleiss, Joseph L.},
    title={Measuring nominal scale agreement among many raters},
    journal={Psychological Bulletin},
    year={1971},
    publisher={American Psychological Association},
    address={US},
    volume={76},
    number={5},
    pages={378-382},
    doi={10.1037/h0031619},
    url={https://doi.org/10.1037/h0031619}
}

@inproceedings{guo-etal-2025-care,
    title = "{CARE}: Multilingual Human Preference Learning for Cultural Awareness",
    author = "Guo, Geyang  and
      Naous, Tarek  and
      Wakaki, Hiromi  and
      Nishimura, Yukiko  and
      Mitsufuji, Yuki  and
      Ritter, Alan  and
      Xu, Wei",
    editor = "Christodoulopoulos, Christos  and
      Chakraborty, Tanmoy  and
      Rose, Carolyn  and
      Peng, Violet",
    booktitle = "Proceedings of the 2025 Conference on Empirical Methods in Natural Language Processing",
    month = nov,
    year = "2025",
    address = "Suzhou, China",
    publisher = "Association for Computational Linguistics",
    url = "https://aclanthology.org/2025.emnlp-main.1669/",
    doi = "10.18653/v1/2025.emnlp-main.1669",
    pages = "32854--32883",
    ISBN = "979-8-89176-332-6",
}

@inproceedings{laiyk-etal-2025-instruction,
    title = "Instruction Tuning on Public Government and Cultural Data for Low-Resource Language: a Case Study in {K}azakh",
    author = "Laiyk, Nurkhan  and
      Orel, Daniil  and
      Joshi, Rituraj  and
      Goloburda, Maiya  and
      Wang, Yuxia  and
      Nakov, Preslav  and
      Koto, Fajri",
    editor = "Che, Wanxiang  and
      Nabende, Joyce  and
      Shutova, Ekaterina  and
      Pilehvar, Mohammad Taher",
    booktitle = "Proceedings of the 63rd Annual Meeting of the Association for Computational Linguistics (Volume 1: Long Papers)",
    month = jul,
    year = "2025",
    address = "Vienna, Austria",
    publisher = "Association for Computational Linguistics",
    url = "https://aclanthology.org/2025.acl-long.706/",
    doi = "10.18653/v1/2025.acl-long.706",
    pages = "14509--14538",
    ISBN = "979-8-89176-251-0",
}

@article{li2023bactrianx,
    title={Bactrian-{X}: Multilingual Replicable Instruction-Following Models with Low-Rank Adaptation}, 
    author={Haonan Li and Fajri Koto and Minghao Wu and Alham Fikri Aji and Timothy Baldwin},
    year={2023},
    journal={arXiv preprint arXiv:2305.15011},
    url={https://arxiv.org/abs/2305.15011}, 
}

@inproceedings{rachamalla-etal-2025-pragyaan,
    title = "Pragyaan: Designing and Curating High-Quality Cultural Post-Training Datasets for {I}ndian Languages",
    author = "Rachamalla, Neel Prabhanjan  and
      Konakalla, Aravind  and
      Rajeev, Gautam  and
      Kulkarni, Ashish  and
      Khatri, Chandra  and
      Agarwal, Shubham",
    editor = "Adelani, David Ifeoluwa  and
      Arnett, Catherine  and
      Ataman, Duygu  and
      Chang, Tyler A.  and
      Gonen, Hila  and
      Raja, Rahul  and
      Schmidt, Fabian  and
      Stap, David  and
      Wang, Jiayi",
    booktitle = "Proceedings of the 5th Workshop on Multilingual Representation Learning (MRL 2025)",
    month = nov,
    year = "2025",
    address = "Suzhuo, China",
    publisher = "Association for Computational Linguistics",
    url = "https://aclanthology.org/2025.mrl-main.20/",
    doi = "10.18653/v1/2025.mrl-main.20",
    pages = "285--321",
    ISBN = "979-8-89176-345-6",
}

@misc{openai2025gpt5,
    title={{GPT-5 System Card}},
    author={OpenAI},
    year={2025},
    url={https://cdn.openai.com/papers/gpt-5-system-card.pdf},
}

@article{lee2024presence,
  title={Presence of social minorities in high school social studies textbooks in the post-democratic era in {S}outh {K}orea},
  author={Lee, Jungwoo and Eun, Ji-Yong},
  journal={Cogent Education},
  volume={11},
  number={1},
  pages={2406580},
  year={2024},
  publisher = {Cogent OA},
  doi = {10.1080/2331186X.2024.2406580},
  URL = {https://doi.org/10.1080/2331186X.2024.2406580},
  eprint = {https://doi.org/10.1080/2331186X.2024.2406580}
}

@article{cho2023analysis,
  title={Analysis of the Issues That Emerged in the Revision of the National Social Studies Curriculum in {S}outh {K}orea: Text Mining and Semantic Network Analysis of the Comments at the Public Hearing on {Y}ou{T}ube.},
  author={Cho, Chul-Ki and Kim, HyeSook and Lee, Soyoung},
  journal={Journal of Education and E-Learning Research},
  volume={10},
  number={3},
  pages={463--473},
  year={2023},
  publisher={ERIC},
  doi={10.20448/jeelr.v10i3.4886}
}

@inproceedings{cho-etal-2025-hermit,
    title = "Hermit Kingdom Through the Lens of Multiple Perspectives: A Case Study of {LLM} Hallucination on {N}orth {K}orea",
    author = "Cho, Eunjung  and
      Cho, Won Ik  and
      Seo, Soomin",
    editor = "Rambow, Owen  and
      Wanner, Leo  and
      Apidianaki, Marianna  and
      Al-Khalifa, Hend  and
      Eugenio, Barbara Di  and
      Schockaert, Steven",
    booktitle = "Proceedings of the 31st International Conference on Computational Linguistics",
    month = jan,
    year = "2025",
    address = "Abu Dhabi, UAE",
    publisher = "Association for Computational Linguistics",
    url = "https://aclanthology.org/2025.coling-main.226/",
    pages = "3353--3371",
}

@Article{cohen1960coefficient,
    author={Cohen, Jacob},
    title={A coefficient of agreement for nominal scales},
    journal={Educational and Psychological Measurement},
    year={1960},
    publisher={Sage Publications},
    address={US},
    volume={20},
    pages={37-46},
    doi={10.1177/001316446002000104},
    url={https://doi.org/10.1177/001316446002000104}
}

@inproceedings{shi-etal-2024-culturebank,
    title = "{C}ulture{B}ank: An Online Community-Driven Knowledge Base Towards Culturally Aware Language Technologies",
    author = "Shi, Weiyan  and
      Li, Ryan  and
      Zhang, Yutong  and
      Ziems, Caleb  and
      Yu, Sunny  and
      Horesh, Raya  and
      Paula, Rog{\'e}rio Abreu De  and
      Yang, Diyi",
    editor = "Al-Onaizan, Yaser  and
      Bansal, Mohit  and
      Chen, Yun-Nung",
    booktitle = "Findings of the Association for Computational Linguistics: EMNLP 2024",
    month = nov,
    year = "2024",
    address = "Miami, Florida, USA",
    publisher = "Association for Computational Linguistics",
    url = "https://aclanthology.org/2024.findings-emnlp.288/",
    doi = "10.18653/v1/2024.findings-emnlp.288",
    pages = "4996--5025",
}

@inproceedings{kim-etal-2024-click,
    title = "{CLI}c{K}: A Benchmark Dataset of Cultural and Linguistic Intelligence in {K}orean",
    author = "Kim, Eunsu  and
      Suk, Juyoung  and
      Oh, Philhoon  and
      Yoo, Haneul  and
      Thorne, James  and
      Oh, Alice",
    editor = "Calzolari, Nicoletta  and
      Kan, Min-Yen  and
      Hoste, Veronique  and
      Lenci, Alessandro  and
      Sakti, Sakriani  and
      Xue, Nianwen",
    booktitle = "Proceedings of the 2024 Joint International Conference on Computational Linguistics, Language Resources and Evaluation (LREC-COLING 2024)",
    month = may,
    year = "2024",
    address = "Torino, Italia",
    publisher = "ELRA and ICCL",
    url = "https://aclanthology.org/2024.lrec-main.296/",
    pages = "3335--3346",
}

@inproceedings{myung2024blend,
 author = {Myung, Junho and Lee, Nayeon and Zhou, Yi and Jin, Jiho and Putri, Rifki Afina and Antypas, Dimosthenis and Borkakoty, Hsuvas and Kim, Eunsu and Perez-Almendros, Carla and Ayele, Abinew Ali and Guti\'{e}rrez-Basulto, V\'{\i}ctor and Ib\'{a}\~{n}ez-Garc\'{\i}a, Yazm\'{\i}n and Lee, Hwaran and Muhammad, Shamsuddeen Hassan and Park, Kiwoong and Rzayev, Anar Sabuhi and White, Nina and Yimam, Seid Muhie and Pilehvar, Mohammad Taher and Ousidhoum, Nedjma and Camacho-Collados, Jose and Oh, Alice},
 booktitle = {Advances in Neural Information Processing Systems},
 doi = {10.52202/079017-2483},
 editor = {A. Globerson and L. Mackey and D. Belgrave and A. Fan and U. Paquet and J. Tomczak and C. Zhang},
 pages = {78104--78146},
 publisher = {Curran Associates, Inc.},
 title = {{BLE}n{D}: A Benchmark for LLMs on Everyday Knowledge in Diverse Cultures and Languages},
 url = {https://proceedings.neurips.cc/paper_files/paper/2024/file/8eb88844dafefa92a26aaec9f3acad93-Paper-Datasets_and_Benchmarks_Track.pdf},
 volume = {37},
 year = {2024}
}

@inproceedings{chiu-etal-2025-culturalbench,
    title = "{C}ultural{B}ench: A Robust, Diverse and Challenging Benchmark for Measuring {LM}s' Cultural Knowledge Through Human-{AI} Red-Teaming",
    author = "Chiu, Yu Ying  and
      Jiang, Liwei  and
      Lin, Bill Yuchen  and
      Park, Chan Young  and
      Li, Shuyue Stella  and
      Ravi, Sahithya  and
      Bhatia, Mehar  and
      Antoniak, Maria  and
      Tsvetkov, Yulia  and
      Shwartz, Vered  and
      Choi, Yejin",
    editor = "Che, Wanxiang  and
      Nabende, Joyce  and
      Shutova, Ekaterina  and
      Pilehvar, Mohammad Taher",
    booktitle = "Proceedings of the 63rd Annual Meeting of the Association for Computational Linguistics (Volume 1: Long Papers)",
    month = jul,
    year = "2025",
    address = "Vienna, Austria",
    publisher = "Association for Computational Linguistics",
    url = "https://aclanthology.org/2025.acl-long.1247/",
    doi = "10.18653/v1/2025.acl-long.1247",
    pages = "25663--25701",
    ISBN = "979-8-89176-251-0",
}

@inproceedings{rei-etal-2022-cometkiwi,
    title = "{C}omet{K}iwi: {IST}-Unbabel 2022 Submission for the Quality Estimation Shared Task",
    author = "Rei, Ricardo  and
      Treviso, Marcos  and
      Guerreiro, Nuno M.  and
      Zerva, Chrysoula  and
      Farinha, Ana C  and
      Maroti, Christine  and
      C. de Souza, Jos{\'e} G.  and
      Glushkova, Taisiya  and
      Alves, Duarte  and
      Coheur, Luisa  and
      Lavie, Alon  and
      Martins, Andr{\'e} F. T.",
    editor = {Koehn, Philipp  and
      Barrault, Lo{\"i}c  and
      Bojar, Ond{\v{r}}ej  and
      Bougares, Fethi  and
      Chatterjee, Rajen  and
      Costa-juss{\`a}, Marta R.  and
      Federmann, Christian  and
      Fishel, Mark  and
      Fraser, Alexander  and
      Freitag, Markus  and
      Graham, Yvette  and
      Grundkiewicz, Roman  and
      Guzman, Paco  and
      Haddow, Barry  and
      Huck, Matthias  and
      Jimeno Yepes, Antonio  and
      Kocmi, Tom  and
      Martins, Andr{\'e}  and
      Morishita, Makoto  and
      Monz, Christof  and
      Nagata, Masaaki  and
      Nakazawa, Toshiaki  and
      Negri, Matteo  and
      N{\'e}v{\'e}ol, Aur{\'e}lie  and
      Neves, Mariana  and
      Popel, Martin  and
      Turchi, Marco  and
      Zampieri, Marcos},
    booktitle = "Proceedings of the Seventh Conference on Machine Translation (WMT)",
    month = dec,
    year = "2022",
    address = "Abu Dhabi, United Arab Emirates (Hybrid)",
    publisher = "Association for Computational Linguistics",
    url = "https://aclanthology.org/2022.wmt-1.60/",
    doi = "10.18653/v1/2022.wmt-1.60",
    pages = "634--645",
}

@inproceedings{geva-etal-2019-modeling,
    title = "Are We Modeling the Task or the Annotator? An Investigation of Annotator Bias in Natural Language Understanding Datasets",
    author = "Geva, Mor  and
      Goldberg, Yoav  and
      Berant, Jonathan",
    editor = "Inui, Kentaro  and
      Jiang, Jing  and
      Ng, Vincent  and
      Wan, Xiaojun",
    booktitle = "Proceedings of the 2019 Conference on Empirical Methods in Natural Language Processing and the 9th International Joint Conference on Natural Language Processing (EMNLP-IJCNLP)",
    month = nov,
    year = "2019",
    address = "Hong Kong, China",
    publisher = "Association for Computational Linguistics",
    url = "https://aclanthology.org/D19-1107/",
    doi = "10.18653/v1/D19-1107",
    pages = "1161--1166",
}

@article{paullada2021data,
    author={Paullada, Amandalynne and Raji, Inioluwa Deborah and Bender, Emily M. and Denton, Emily and Hanna, Alex},
    title={Data and its (dis)contents: A survey of dataset development and use in machine learning research},
    journal={Patterns},
    year={2021},
    month={Nov},
    day={12},
    publisher={Elsevier},
    volume={2},
    number={11},
    issn={2666-3899},
    doi={10.1016/j.patter.2021.100336},
    url={https://doi.org/10.1016/j.patter.2021.100336}
}

@inproceedings{cui-etal-2025-bias,
    title = "Bias in, Bias out: Annotation Bias in Multilingual Large Language Models",
    author = "Cui, Xia  and
      Huang, Ziyi  and
      Adel, Naeemeh",
    editor = "Przyby{\l}a, Piotr  and
      Shardlow, Matthew  and
      Colombatto, Clara  and
      Inie, Nanna",
    booktitle = "Proceedings of Interdisciplinary Workshop on Observations of Misunderstood, Misguided and Malicious Use of Language Models",
    month = sep,
    year = "2025",
    address = "Varna, Bulgaria",
    publisher = "INCOMA Ltd., Shoumen, Bulgaria",
    url = "https://aclanthology.org/2025.ommm-1.1/",
    pages = "1--16",
}

@inproceedings{tan-etal-2024-large,
    title = "Large Language Models for Data Annotation and Synthesis: A Survey",
    author = "Tan, Zhen  and
      Li, Dawei  and
      Wang, Song  and
      Beigi, Alimohammad  and
      Jiang, Bohan  and
      Bhattacharjee, Amrita  and
      Karami, Mansooreh  and
      Li, Jundong  and
      Cheng, Lu  and
      Liu, Huan",
    editor = "Al-Onaizan, Yaser  and
      Bansal, Mohit  and
      Chen, Yun-Nung",
    booktitle = "Proceedings of the 2024 Conference on Empirical Methods in Natural Language Processing",
    month = nov,
    year = "2024",
    address = "Miami, Florida, USA",
    publisher = "Association for Computational Linguistics",
    url = "https://aclanthology.org/2024.emnlp-main.54/",
    doi = "10.18653/v1/2024.emnlp-main.54",
    pages = "930--957",
}

@article{gilardi2023chatgpt,
    author = {Fabrizio Gilardi  and Meysam Alizadeh  and Maël Kubli },
    title = {{C}hat{GPT} outperforms crowd workers for text-annotation tasks},
    journal = {Proceedings of the National Academy of Sciences},
    volume = {120},
    number = {30},
    pages = {e2305016120},
    year = {2023},
    doi = {10.1073/pnas.2305016120},
    URL = {https://www.pnas.org/doi/abs/10.1073/pnas.2305016120},
    eprint = {https://www.pnas.org/doi/pdf/10.1073/pnas.2305016120},
}
